%% file: main.tex
\pdfoutput=1
\documentclass{article}

\usepackage{microtype}
\usepackage{graphicx}

\usepackage{booktabs} 

\usepackage{amsmath}
\usepackage{amssymb}
\usepackage{mathtools}
\usepackage{amsthm}
\usepackage{multirow}
\usepackage{colortbl}
\usepackage{pifont}
\usepackage{subcaption}
\usepackage{algorithm, algorithmic}
\usepackage{wrapfig}
\usepackage{todonotes}
\usepackage{hyperref}

\usepackage{amsmath}
\usepackage{amssymb}
\usepackage{mathtools}
\usepackage{amsthm}
\usepackage{multirow}
\usepackage{colortbl}
\usepackage{pifont}

\usepackage{wrapfig}
\usepackage{mwe,graphicx,caption,subcaption}

\usepackage{graphicx}
\usepackage[utf8]{inputenc} 
\usepackage[T1]{fontenc}    
\usepackage{hyperref}       
\usepackage{url}            
\usepackage{booktabs}       
\usepackage{amsfonts}       
\usepackage{nicefrac}       
\usepackage{microtype}      
\usepackage{xcolor}         
\usepackage{graphicx}
\usepackage{hyperref}
\usepackage{inconsolata}
\usepackage[accepted]{mlsys2025}
\usepackage{soul}

\newcommand{\bm}{\boldsymbol}
\newcommand{\R}{\mathbb{R}}
\newcommand{\nn}{n}
\newcommand{\nh}{H}
\newcommand{\nng}{n_{g}}
\newcommand{\nbu}{n_{b}}
\newcommand{\nhu}{n_{h}}
\newcommand{\dd}{d}
\newcommand{\ddh}{d_{H}}
\newcommand{\bb}{b}

\newcommand{\rr}{r}
\newcommand{\gs}{g}
\newcommand{\bx}{\bm{x}}
\newcommand{\bq}{\bm{q}}
\newcommand{\bk}{\bm{k}}
\newcommand{\bv}{\bm{v}}
\newcommand{\bxt}{\bx_{t}}
\newcommand{\bqt}{\bq_{t}}
\newcommand{\bkt}{\bk_{t}}
\newcommand{\bvt}{\bv_{t}}

\newcommand{\mX}{\bm{X}}
\newcommand{\mH}{\bm{H}}
\newcommand{\mW}{\bm{W}}
\newcommand{\mQ}{\bm{Q}}
\newcommand{\mK}{\bm{K}}
\newcommand{\mV}{\bm{V}}

\newcommand{\Wo}{\mW_{o}}

\newcommand{\mQh}{\mQ^{(h)}}
\newcommand{\mKh}{\mK^{(h)}}
\newcommand{\mVh}{\mV^{(h)}}
\newcommand{\Wqh}{\mW_{q_{h}}}
\newcommand{\Wkh}{\mW_{k_{h}}}
\newcommand{\Wvh}{\mW_{v_{h}}}

\newcommand{\Bcal}{\mathcal{B}}

\newcommand{\vQ}{\mathbf{Q}}
\newcommand{\vK}{\mathbf{K}}
\newcommand{\vV}{\mathbf{V}}

\newcommand{\vS}{\mathbf{S}}

\newcommand{\vP}{\mathbf{P}}

\newcommand{\vO}{\mathbf{O}}

\newcommand{\diag}{\mathrm{diag}}

\newcommand{\ourquant}{FlashQ}
\newcommand{\oursoftmax}{SAS}
\newcommand{\ouralg}{TurboAttention}

\mlsystitlerunning{TurboAttention: Efficient attention approximation for High Throughputs LLMs}

\begin{document}

\twocolumn[
\mlsystitle{TurboAttention: Efficient attention approximation for High Throughputs LLMs}



\mlsyssetsymbol{equal}{*}

\begin{mlsysauthorlist}
\mlsysauthor{Hao Kang}{to,goo}
\mlsysauthor{Srikant Bharadwaj}{to}
\mlsysauthor{James Hensman}{to}
\mlsysauthor{Tushar Krishna}{goo}
\mlsysauthor{Victor Rühle}{to}
\mlsysauthor{Saravan Rajmohan}{to}
\end{mlsysauthorlist}

\mlsysaffiliation{to}{Microsoft}
\mlsysaffiliation{goo}{Georgia Institute of Technology}

\mlsyscorrespondingauthor{Srikant Bharadwaj}{srikant.bharadwaj@microsoft.com}

\mlsyskeywords{Machine Learning, MLSys}

\vskip 0.3in

\input{Section/Abstract}
]



\printAffiliationsAndNotice{\mlsysEqualContribution} 

\input{Section/Introduction}
\input{Section/Background}

\input{Section/FlashQuant}  
\input{Section/SAS}

\input{Section/Evaluations}
\input{Section/Conclusions}

\nocite{langley00}

\bibliography{example_paper}
\bibliographystyle{mlsys2024}

\appendix
\input{Section/Appendix}
%


\end{document}

%% file: Section/Abstract.tex
\begin{abstract}

Large language model (LLM) inference demands significant amount of computation and memory, especially in the key attention mechanism. While techniques, such as quantization and acceleration algorithms, like FlashAttention, have improved efficiency of the overall inference, they address different aspects of the problem: quantization focuses on weight-activation operations, while FlashAttention improves execution but requires high-precision formats. Recent Key-value (KV) cache quantization reduces memory bandwidth but still needs floating-point dequantization for attention operation.

We present {\it{\ouralg}}, a comprehensive approach to enable quantized execution of attention that simultaneously addresses both memory and computational efficiency. Our solution introduces two key innovations: {\it{\ourquant}}, a headwise attention quantization technique that enables both compression of KV cache and quantized execution of activation-activation multiplication, and Sparsity-based Softmax Approximation ({\it{\oursoftmax}}), which eliminates the need for dequantization to FP32 during exponentiation operation in attention. Experimental results demonstrate that TurboAttention achieves 1.2-1.8x speedup in attention, reduces the KV cache size by over 4.4x, and enables up to 2.37x maximum throughput over the FP16 baseline while outperforming state-of-the-art quantization and compression techniques across various datasets and models.

\end{abstract}

%% file: Section/Introduction.tex
\section{Introduction}\label{sec:introduction}

\begin{figure*}[t!]  
    \centering
    \begin{subfigure}{0.32\textwidth}
        \centering
        \includegraphics[width=0.86\textwidth]{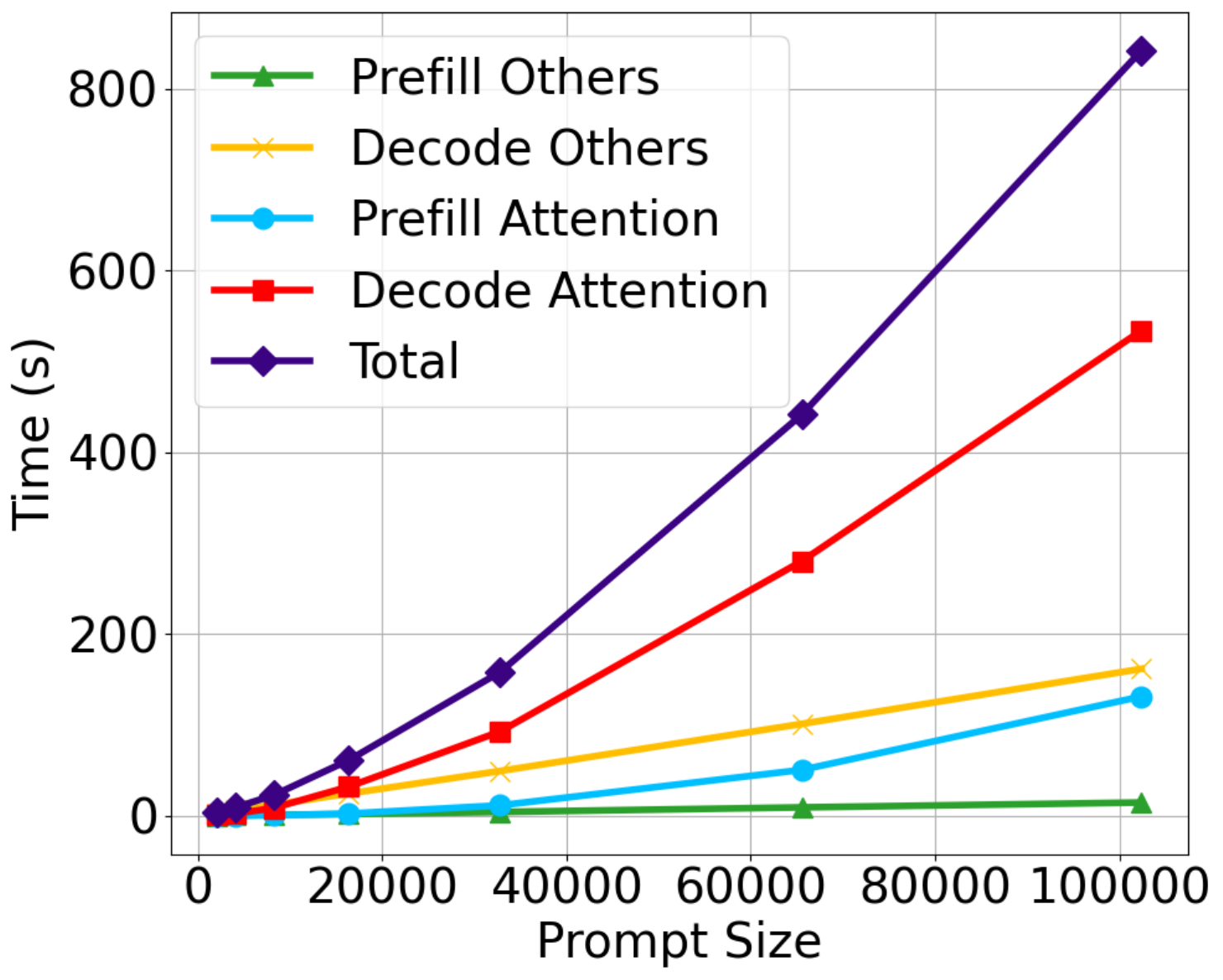}  
        \caption{Prompt:Output Tokens = 8:1}
        \label{fig:context_profile}
    \end{subfigure}
    \begin{subfigure}{0.32\textwidth}
        \centering
        \includegraphics[width=0.86\textwidth]{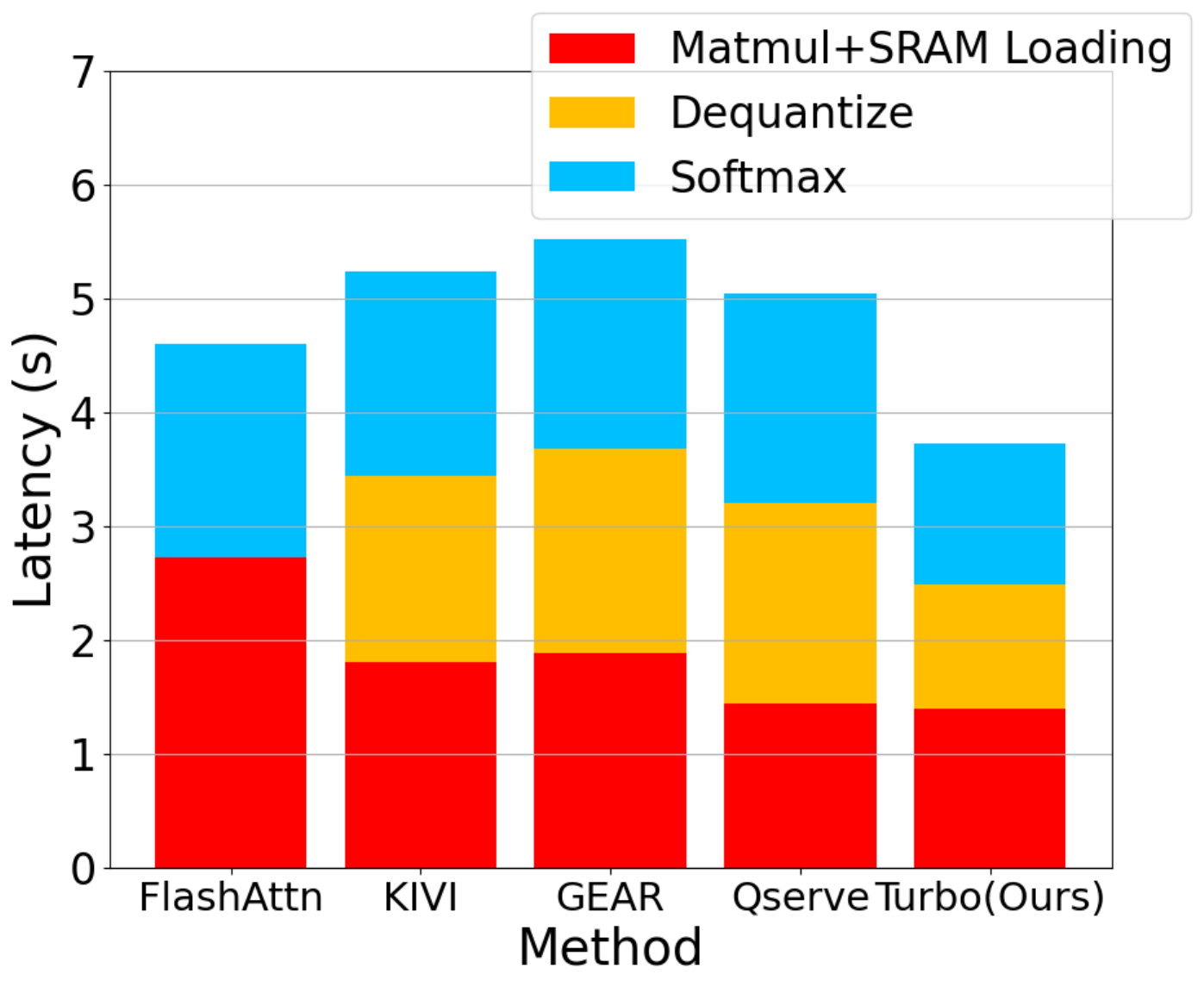}  
        \caption{Timeshare in GPU attention kernel}
        \label{fig:attn_profile}
    \end{subfigure}
    \begin{subfigure}{0.32\textwidth}
        \centering
        \includegraphics[width=0.88\textwidth]{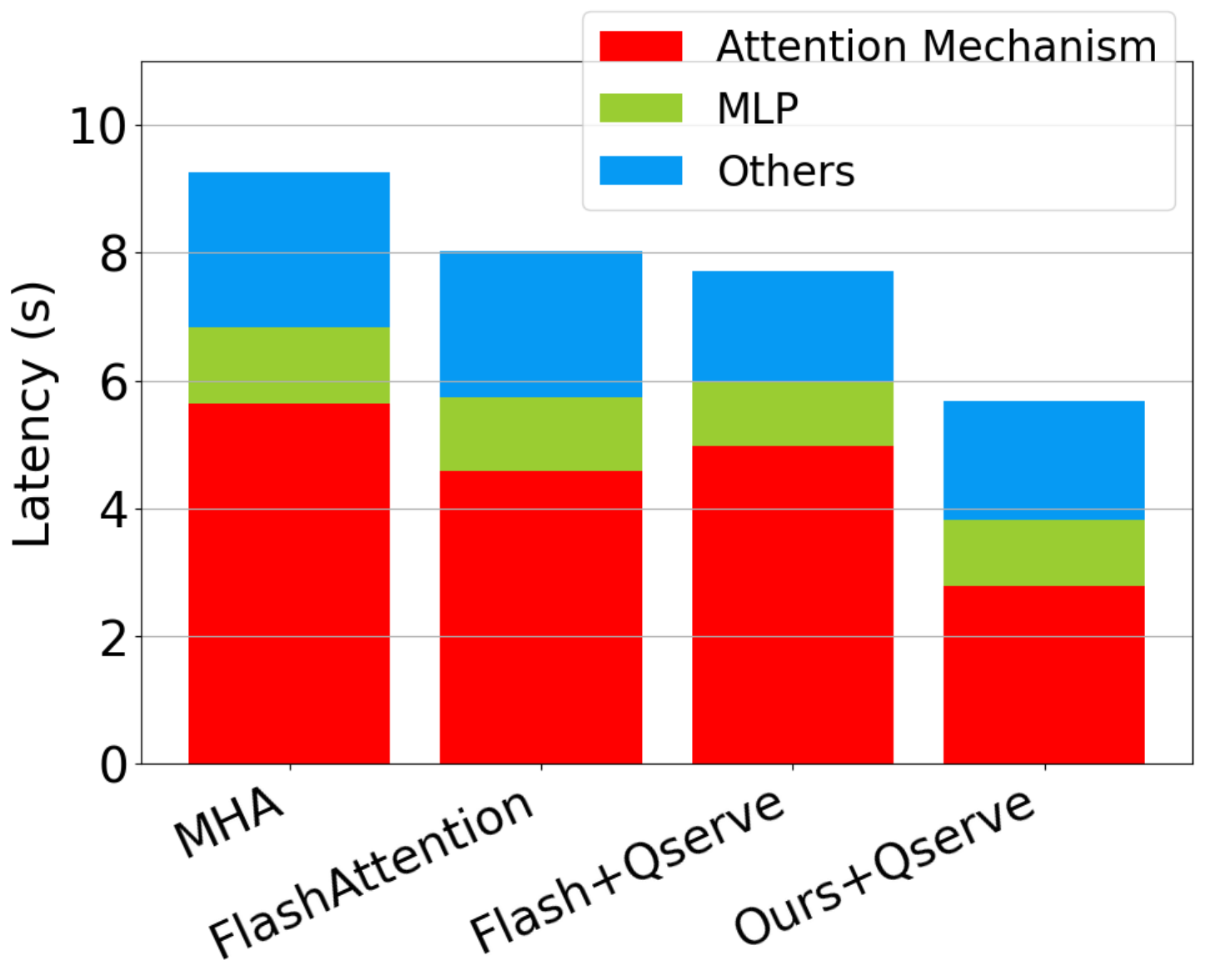}  
        \caption{End-to-end inference timeshare}
        \label{fig:model_profile}
    \end{subfigure}
    \vspace*{-0.2mm}
    \caption{Latency profile of Phi3-Medium on Nvidia A100 GPU. (a) KV cache compression techniques impose a dequantization overhead in attention kernel latency. (b) {\it{\ouralg}} significantly improves attention kernel latency compared to FlashAttention(FP16) baseline while other work mainly focuses on reducing KV-cache memory footprint and bandwidth only.(c) {\it{\ouralg}} reduces latency of Matmul+KV-cache load by enabling quantized integer inference, dequantization by applying block progressive quantization, and faster softmax by introducing sparse activated softmax.}
    \vspace{-4mm}
    \label{fig:three_subfigures}
\end{figure*}

Large language models (LLMs) \citep{touvron2023llama,gunasekar2023textbooksneed,brown2020languagemodelsfewshotlearners} have excelled in tasks like natural language understanding \citep{joshi2017triviaqa,dodge2021c4} and generative text production \citep{hendrycks2021math,zhongSeq2SQL2017}. However, as model size increases, computational and memory demands scale correspondingly, particularly during inference. This necessitates efficient strategies to reduce memory utilization and computational complexity, thereby reducing inference latency and improving throughput — critical requirements for real-time applications that demand faster user experiences.


Quantization is among the most commonly used optimization techniques that simultaneously addresses both computational and memory constraints in LLM inference. By reducing the numerical precision of model parameters, KV cache states, and activation values, this technique enables low-precision forward pass computations while significantly reducing memory footprint. This unified approach to precision reduction offers a systematic method for optimizing inference efficiency when applied to the computational and memory intensive parts of the LLM inference.

The bottlenecks during LLM inference can be split into three major sections: the linear projection operations (QKV projection and FFN), the memory-intensive Key/Value (KV) cache operations, and the attention mechanism's square computational and memory complexity with respect to context length. As both context lengths and model sizes scale up, the KV cache footprint expands substantially, while the attention computations become increasingly resource-intensive, leading to reduced throughput and elevated latencies, which ultimately results in degraded user experiences as well as increased costs. \autoref{fig:context_profile} shows the time spent in the attention operation compared to others during inference as prompt sizes increases. The attention mechanism contributes significantly to the overall generation time, contributing up to 80\% of the overall latency at large context lengths (>80k).

Previous quantization works, such as Atom \citep{zhao2024atom}, QuaRot\citep{ashkboos2024quarot}, and Qserve \citep{lin2024qserve}, have primarily applied quantization techniques to linear network components (such as QKV and FFN projections), converting floating-point parameters and activations to low-bit integers to enhance memory efficiency and computational performance. 
On the other hand, approaches such as KIVI \citep{KIVI}, GEAR \citep{kang2024gear}, and Qserve \citep{lin2024qserve} focus on compressing the KV cache to 4-bit or even 2-bit formats but rely on time-intensive floating-point decompression before executing the attention mechanism. This incurs decompression overhead during the execution of attention leading to increased overall latencies, as shown in \autoref{fig:attn_profile}.
In general, these approaches largely neglect the latency overhead of the attention operation, which is a key bottleneck in terms of latency overhead during inference.




On the other hand, attention execution acceleration methods, such as Flash Attention-1,2,3 \citep{dao2022flashatten, dao2023flashattention2, shah2024flashattention3}, along with related optimizations \citep{flashdecoding, hong2024flashdecoding}, have improved efficiency of the attention operation and have become the commercial standard. However, they operate exclusively on high-precision formats (FP16/32) which leads to high attention latencies at longer context lengths, as shown in \autoref{fig:model_profile}. This leaves considerable scope for further optimizations using quantization, which could yield substantial benefits in memory usage and computational speed.


In this paper, we present, {\it{\ouralg}},  a novel unified technique for enabling quantized execution of attention along with a cooperative KV cache compression mechanism which reduces latency, memory footprint, incurs negligible accuracy loss. First, to accelerate the matrix multiplications, minimize decompression time, and ensure compatibility with FlashAttention, we introduce {\it{\ourquant}} (\autoref{sec:flashq}). Second, for efficient GPU tensor core-friendly, softmax computation, we develop {\it{\oursoftmax}} (\autoref{sec:sas}). Third, to alleviate the memory bottleneck posed by the KV cache, we design a CUDA kernel-friendly headwise mixed precision quantization method(\autoref{sec:flashq}). Together, {\it{\ourquant}} and {\it{\oursoftmax}} form our comprehensive solution, {\it{\ouralg}}. To our knowledge, this is the \ul{first work to bridge the state-of-the-art attention acceleration method (FlashAttention) with a quantization algorithm}, and the first to apply lossy techniques to accelerate the entire attention mechanism, encompassing both matrix multiplication and softmax operations. A comparison of our approach with concurrent works is provided in \autoref{tab:feature_comparison}.

Our evaluations show that {\it{\ouralg}} outperforms state-of-the-art FlashAttention-3 on delivering up to 1.8x improvement in latency as well as up to $2.37\times$ maximum throughput. Further, our method delivers near-lossless accuracy across various tasks including mathematical reasoning 
(GSM8k and AQuA), symbolic reasoning (BigBench-Hard) and models (Llama3, Qwen2 and Phi-3), further validating its effectiveness.

\begin{table*}[]
\centering
\caption{TurboAttention enables quantized execution of attention operation as well as KV cache compression. Note that techniques such as ATOM\cite{zhao2024atom}, QuaRot\cite{ashkboos2024quarot} are orthogonal to TurboAttention and can easily be applied in conjunction.}
\label{tab:feature_comparison}
\resizebox{\textwidth}{!}{%
\begin{tabular}{
>{\columncolor[HTML]{FFFFFF}}c |
>{\columncolor[HTML]{FFFFFF}}l |
>{\columncolor[HTML]{FFFFFF}}c |
>{\columncolor[HTML]{FFFFFF}}c |
>{\columncolor[HTML]{FFFFFF}}c |
>{\columncolor[HTML]{FFFFFF}}c |
>{\columncolor[HTML]{EFEFEF}}c |
>{\columncolor[HTML]{EFEFEF}}c |}
\cline{2-8}
\cellcolor[HTML]{343434}{\color[HTML]{FFFFFF} \textbf{Target}} &
  \multicolumn{1}{c|}{\cellcolor[HTML]{343434}{\color[HTML]{FFFFFF} \textbf{Technique}}} &
  \cellcolor[HTML]{343434}{\color[HTML]{FFFFFF} \textbf{\begin{tabular}[c]{@{}c@{}}QKV\\ Projection\end{tabular}}} &
  \cellcolor[HTML]{343434}{\color[HTML]{FFFFFF} \textbf{\begin{tabular}[c]{@{}c@{}}KV Cache\\ Compression\end{tabular}}} &
  \cellcolor[HTML]{343434}{\color[HTML]{FFFFFF} \textbf{\begin{tabular}[c]{@{}c@{}}Attention\\ Execution\end{tabular}}} &
  \cellcolor[HTML]{343434}{\color[HTML]{FFFFFF} \textbf{MLP}} &
  \cellcolor[HTML]{343434}{\color[HTML]{FFFFFF} \textbf{\begin{tabular}[c]{@{}c@{}}Memory \\ Overhead\end{tabular}}} &
  \cellcolor[HTML]{343434}{\color[HTML]{FFFFFF} \textbf{\begin{tabular}[c]{@{}c@{}}Inference\\  Latency\end{tabular}}} \\ \cline{2-8} 
\multicolumn{1}{|c|}{\cellcolor[HTML]{FFFFFF}} &
  ATOM \cite{zhao2024atom} &
  Quantized &
  $\textcolor[rgb]{0,0.5,0}{\checkmark}$ &
  - &
  Quantized &
  {$\textcolor[rgb]{0,0.5,0}{\downarrow}$} &
  {$\textcolor[rgb]{0,0.5,0}{\downarrow}$} \\ \cline{2-8} 
\multicolumn{1}{|c|}{\cellcolor[HTML]{FFFFFF}} &
  QuaRot \cite{ashkboos2024quarot} &
  Quantized &
  $\textcolor[rgb]{0,0.5,0}{\checkmark}$ &
  - &
  Quantized &
  {$\textcolor[rgb]{0,0.5,0}{\downarrow}$} &
  {$\textcolor[rgb]{0,0.5,0}{\downarrow}$} \\ \cline{2-8} 
\multicolumn{1}{|c|}{\multirow{-3}{*}{\cellcolor[HTML]{FFFFFF}\begin{tabular}[c]{@{}c@{}}Linear Operation\\ (QKV, MLP, etc.)\end{tabular}}} &
  Qserve \cite{lin2024qserve} &
  Quantized &
  {$\textcolor[rgb]{0,0.5,0}{\checkmark}$} &
  - &
  Quantized &
  {$\textcolor[rgb]{0,0.5,0}{\downarrow \downarrow}$} &
  {$\textcolor[rgb]{0,0.5,0}{\downarrow \downarrow}$} \\ \hline
\multicolumn{1}{|c|}{\cellcolor[HTML]{FFFFFF}} &
  KIVI \cite{KIVI} &
  - &
  {$\textcolor[rgb]{0,0.5,0}{\checkmark}$} &
  - &
  - &
  {$\textcolor[rgb]{0,0.5,0}{\downarrow }$} &
  {$\textcolor[rgb]{0,0.5,0}{\downarrow}$} \\ \cline{2-8} 
\multicolumn{1}{|c|}{\cellcolor[HTML]{FFFFFF}} &
  GEAR \cite{kang2024gear} &
  - &
  {$\textcolor[rgb]{0,0.5,0}{\checkmark}$} &
  - &
  - &
  {$\textcolor[rgb]{0,0.5,0}{\downarrow}$} &
  {$\textcolor[rgb]{0,0.5,0}{\downarrow}$} \\ \cline{2-8} 
\multicolumn{1}{|c|}{\cellcolor[HTML]{FFFFFF}Attention Operation} &
  FlashAttention \cite{dao2022flashatten} &
  - &
  - &
  Flash &
  - &
  {$\textcolor{red}{\times}$} &
  {$\textcolor[rgb]{0,0.5,0}{\downarrow}$} \\ \cline{2-8} 
\multicolumn{1}{|c|}{\cellcolor[HTML]{FFFFFF}} &
  {\bf{\it{\ouralg}}} (This Work) &
  - &
  {$\textcolor[rgb]{0,0.5,0}{\checkmark}$} &
  Flash + Quantized &
  - &
  {$\textcolor[rgb]{0,0.5,0}{\downarrow \downarrow}$} &
  {$\textcolor[rgb]{0,0.5,0}{\downarrow \downarrow}$} \\ \hline
\end{tabular}%
}
\end{table*}

%% file: Section/Background.tex
\section{Background and Motivation}

In this section, we first introduce the key aspects of the attention mechanism and state-of-the-art quantization techniques before discussing the motivation and challenges in enabling a quantized execution of attention mechanism.

\begin{figure*}[t!] 
    \centering
        \centering
        \includegraphics[width=\textwidth]{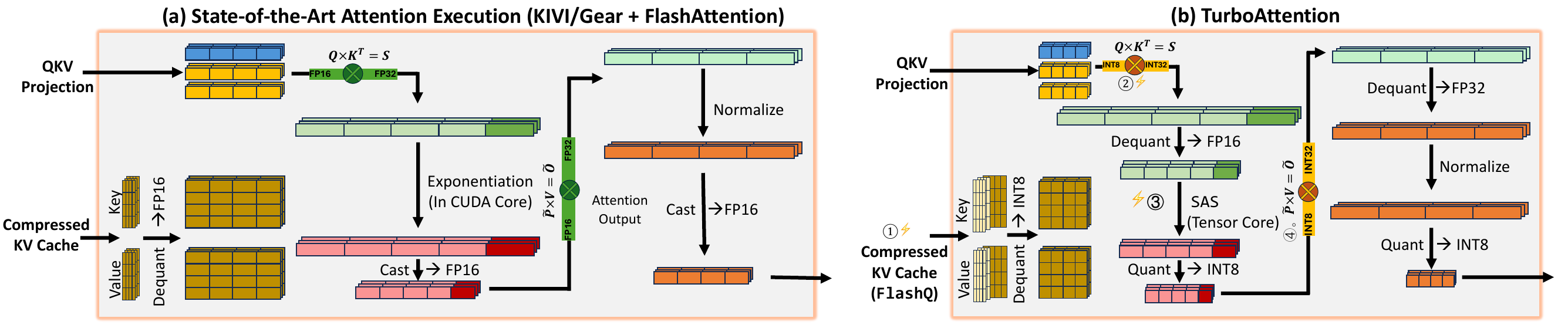}
    \vspace{-8mm}
    \caption{High-Level comparison of {\it{\ouralg}} compared to state-of-the-art KV-cache compression technique combined with FlashAttention. {\it{\ouralg}} accelerates the attention mechanism by adapting (1) FlashQ which enables KV cache compression and accelerated Matmuls ((2) and (3)) and (4) SAS which enables techniques which allow faster execution of attention by utilizing the tensor cores of GPUs efficiently.}
    \label{fig:highlevelflow}
    \vspace{-6mm}
\end{figure*}

\subsection{Attention Mechanism}

\label{sec:attn_mechanism}
Multi-head attention (MHA) is a core component of the transformer architecture, with each transformer model consisting of $L$ stacked layers. Each layer comprises two submodules: a multi-head attention mechanism (MHA) and a feed-forward network (FFN). Given input token embeddings $\mX \in \R^{\nn \times \dd}$, the MHA computes attention in parallel across $\nh$ attention heads, defined as:

\begin{equation}
\begin{aligned}
    \text{MHA}\left(\mX \right) &= \text{Concat}\left(\mH^{(1)}, ..., \mH^{(\nh)}\right) \Wo, \\
    \mH^{(h)} &= \text{Softmax}\left( \frac{\mQh \mK^{(h)\top}}{\sqrt{\ddh}} \right) \mVh
\end{aligned}
\label{eqn:stdattn}
\end{equation}

where $ \mQh = \mX\Wqh, \mKh = \mX\Wkh, \mVh = \mX\Wvh $ are the Query, Key, and Value matrices for head $h$. The projection matrices $ \Wqh, \Wkh, \Wvh \in \R^{\dd\times \ddh} $ map the input embedding $\mX$ to the respective attention components, and $\ddh$ is typically set to $\dd/\nh$ to balance the dimensionality across heads.

\textbf{Prefill and decoding.} During generation, let the model generate $\nng$ tokens. In the initial step, the input tokens $\mX_{0} \in \R^{\nn \times \dd}$ are prefilled, and the Keys and Values ($\mK, \mV$) for each head and each layer are cached for future use. This prefill stage results in the KV caches: $\mK_{0} = \textrm{Concat}(\mK^{(1)}, \dots, \mK^{(\nh)})$ and $\mV_{0} = \textrm{Concat}(\mV^{(1)}, \dots, \mV^{(\nh)})$, where $\mK_{0}, \mV_{0} \in \R^{n \times \dd}$.

At each step $t$ ($1 \leq t \leq \nng$) of autoregressive decoding, the model generates a new token $\bxt$, conditioned on the input and previously generated tokens. For the subsequent steps, multi-head attention (MHA) only computes the Query/Key/Value vectors ($\bqt, \bkt, \bvt \in \R^{\dd}$) for the newly generated token $\bxt$. These vectors are then appended to the KV cache: $\mK_{t} = \mK_{t-1} | \bkt$ and $\mV_{t} = \mV_{t-1} | \bvt$. The attention mechanism is then performed between $\bqt$ and the updated KV cache, i.e., $\mK_{t}$ and $\mV_{t}$.

\subsection{Attention Acceleration Techniques}
\label{sec:attn_acceleration_mechanism}
The query matrix $Q \in R^{N_q \times d}$ and key  and value matrices $K$, $V$ $\in R^{N_k \times d}$ are inputs to the following equation which is computed independently for each head and batch instance. The output matrix $H \in R^{N_q \times d}$ is obtained in essentially three steps as shown in Equation \ref{eqn:threesteps}. 
%
\begin{equation}
    S = QK^\top,\quad\text{ } P = \operatorname{softmax}\left(\frac{S}{\sqrt{d}}\right),\quad\text{ } H = PV\,.
    \label{eqn:threesteps}
\end{equation}
These intermediate activations $S$ and $P$ are large and thus place a significant demand on memory bandwidth. FlashAttention \citep{dao2022flashatten} addresses this by chunking the Query, Key, and Value matrices along the token dimension and tiling the attention mechanism to accelerate the process as described below. 

\label{subsec:FA2background}
\textbf{Flash Attention.} FlashAttention integrates the online softmax algorithm \cite{milakov2018online} to fuse the three operations illustrated in \autoref{eqn:threesteps}. It requires only a single pass over an entire row of tokens to compute attention. This method partitions the attention output matrix $H$ into independent output tiles, with each tile's computation being independent of the others. This approach eliminates the need for intermediate global memory reads and writes of $S$ and $P$ tensors. FlashAttention employs a tiling strategy, where small chunks of $query$, $key$, and $value$ are transferred to the compute cores to first perform ``partial attention'' for each pair of chunks. This is followed by normalization at the end with the help of additional values (sum and max). 

Additionally, optimization techniques such as Ring Attention \cite{liu2023ring}, Striped Attention \cite{brandon2023striped}, and Lean Attention \cite{sanovar2024leanatten} aim to balance computation across GPU resources while using FlashAttention's core algorithm. These techniques significantly reduce the memory bandwidth overhead during inference, especially for long-context generations. However, the tiled dataflow of FlashAttention's core algorithm, as shown in \autoref{fig:highlevelflow}, involving three activation tensors, makes it unsuitable for direct application of state-of-the-art quantization techniques. The quantization techniques do not address the tiled nature of $query$, $key$, and $value$ tensors, thus, techniques which utilize token-wise or channel-wise quantization scale and min cannot be implemented directly without loss of efficiency. On the other hand, these quantization techniques can be applied ot vanilla attention algorithm, but they have the overhead of storing the large intermediate tensors resulting in high memory bandwidth demand.

Further, to avoid overflow, the exponentiation (in partial-attention calculation) in FlashAttention is performed in FP32 data format (exponentiation in \autoref{fig:highlevelflow}(a)), while the matrix multiplications are performed in FP16 using GPU tensor cores. Current generation GPUs are inefficient in handling FP32 and do not make use of tensor cores, resulting in higher latency and inefficiency in the attention operation (FP32 CUDA cores deliver only $\sim$3\% of the performance of FP16 Tensor Cores). The usage of FP16 and FP32 also adds pressure to the scratchpad and register file on the compute units of GPU.

\subsection{Progressive Quantization}
\label{sec:progressive_quantization}
Progressive quantization (PQ) uses a combination of INT8 and INT4 representations progressively, combining the computational efficiency of symmetric quantization with the ability of asymmetric quantization to represent values accurately at INT4. An overview of PQ follows. 

Quantization maps high-precision floating-point values ($x$) into discrete levels ($Q(x)$), expressed as:
\begin{equation}
Q(x) = \left \lceil \frac{x - z}{s} \right \rfloor, \quad \hat{x} = Q(x) \cdot s + z\,,
\end{equation}
where $s$ is a scaling factor and $z$ is a zero-point. These depend on the quantization type, symmetric (sym.) or asymmetric (asym.):
\begin{equation}
s =
\begin{cases}
    \displaystyle \large \frac{x_{\operatorname{max}} - x_{\operatorname{min}}}{2^{\text{bit}} - 1}, & \text{sym.} \\
    \displaystyle \large \frac{\max\left|x \right|}{2^{\text{bit} - 1} - 2}, & \text{asym.}
\end{cases}, \quad
z =
\begin{cases}
    \displaystyle \large 0, & \text{sym.} \\
    \displaystyle \large x_{\text{min}}, & \text{asym.}
\end{cases}
\end{equation}

For two quantized matrices $\hat{A}$ and $\hat{B}$, the low-bit integer matrix multiplication is:
\begin{equation}
\label{fomula:integer_inference}
\begin{aligned}
O_{ij} &\approx \sum_{k} \hat{A_{ik}}\hat{B_{kj}}= s_a s_b \sum_{k} Q(A_{ik}) Q(B_{kj})\\
       & + s_a z_b \sum_{k} Q(A_{ik}) + s_b z_a \sum_{k} Q(B_{kj}) + z_a z_b\,.
\end{aligned}
\end{equation}
We see that asymmetric quantization introduces computation (the last three terms) that are not necessary in symmetric computation (where $z=0$). In general, asymetric quantization leads to lower error rates but results in higher computational overhead.

Progressive quantization proposes two stage of quantization to solve this problem. First, symmetric quantization is used to build INT8 representations for computational efficiency, avoiding the extra overhead of asymmetric quantization (the three non-zero terms above). Then, for better memory efficiency, 8-bit integers are further compressed to INT4 using asymmetric quantization. 

The integer inference formula for PQ is:
\begin{equation}
O_{ij} = s_a s_b \sum_{k} \hat{Q}({A_{ik}}) \hat{Q}({B_{kj}})\,,
\end{equation}
\vspace*{-2mm}
where $\hat{Q}(A)$ and $\hat{Q}(B)$ denote:
\begin{align}
\hat{Q}(A) &= Q(Q(x)) \cdot s_{a}^{\operatorname{int}} + z_{a}^{\operatorname{int}} \label{eq:pq-dequant1}\,,\\
\hat{Q}(B) &= Q(Q(x)) \cdot s_{b}^{\operatorname{int}} + z_{b}^{\operatorname{int}} \,.\label{eq:pq-dequant2}
\end{align}
Here, $s_a$ and $s_b$ are the (FP16) scales from the first INT8 quantization step. Equations \ref{eq:pq-dequant1} and \ref{eq:pq-dequant2} represent dequantization from \emph{asymmetric} INT4 quantization into INT8.
The scales $s^{\operatorname{int}}$ and zero points $z^{\operatorname{int}}$ are stored in INT8, but the majority of the data $Q(Q(x))$ are stored in INT4.  

This approach, proposed in Qserve \citep{lin2024qserve} for weight quantization, is more hardware-friendly compared to decompression or other low-bit integer inference methods, such as those in Atom \citep{zhao2024atom}, LLM.int8 \citep{dettmers2022llmint8}, or TensorRT-LLM .
In our work, we extend the design of PQ to represent KV cache, with the objective of making the dequantization into the attention computations faster while keeping it compatible to attention acceleration mechanisms (such as FlashAttention). 

\begin{figure*}[t]

\begin{center}
\centerline{\includegraphics[width=2.1\columnwidth]{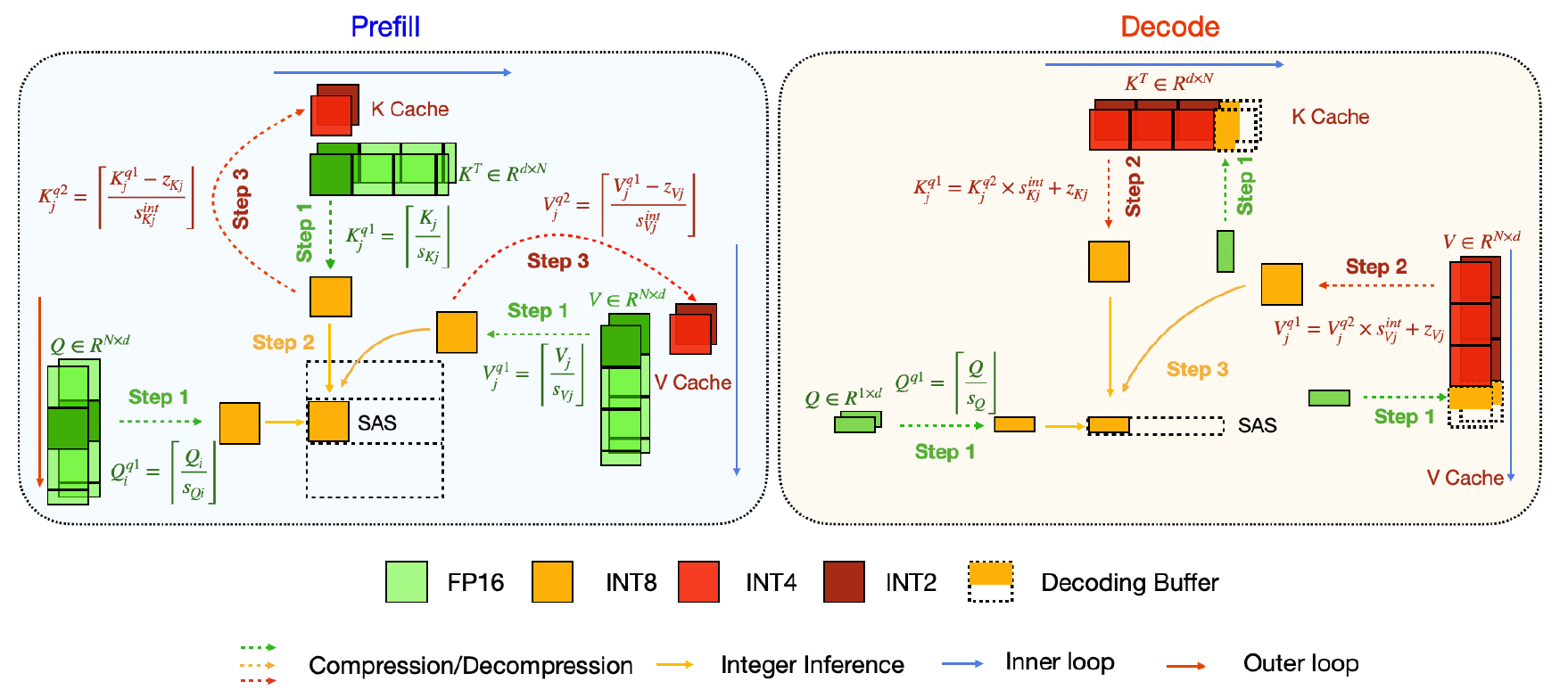}}
\vspace{-4mm}
\caption{Dataflow of {\it{\ouralg}} in pre-fill and decode. At pre-fill (left) we first compress QKV block-wise into INT8(Step1), and compute the attention matrix (on-line) using {\it{\oursoftmax}} (see section 4, Step2). Next, we compress the INT8 KV blocks into asymmetric INT4/INT2, channel-wise, in integer arithmetic: these are stored in the cache(Step3). At decoding (right), we first compress generated qkv to INT8(Step1) and decompress the KV cache to INT8 for integer inference(Step2). Again, we use {\it{\oursoftmax}} to compute attention(Step3).}
\label{fig:workflow}
\vspace{-10mm}
\end{center}

\end{figure*}

\subsection{Challenges in Efficient Attention Mechanism}
\label{sec:motivation}
As we saw in \autoref{fig:context_profile}, the attention operation can constitute up to 80\% of the overall inference execution time when combined with state-of-the-art quantization techniques for other parts of transformer. This is majorly due to the high-precision execution of attention operation in state-of-the-art techniques, such as FlashAttention. 
As we saw in \autoref{sec:attn_acceleration_mechanism}, the usage of FP16 and FP32 not only adds pressure to the scratchpad and register file on the compute units of GPU, but slows down the matrix multiplications and softmax operation inside the attention operation because of the higher precision. For example, the peak performance of FP32 CUDA cores used in FlashAttention's exponentiation operation is only $\sim$3\% of the FP16 tensor performance delivered by Nvidia's A100-SXM. Thus, there is a need to enable quantized execution of the attention operation to leverage the faster low-precision tensor cores in modern GPUs. However, a naive quantization of the attention operation could lead to a loss of the model's generative performance (accuracy).

On the other hand, compression techniques such as KIVI \citep{KIVI}, GEAR \citep{kang2024gear}, and Qserve \citep{lin2024qserve} have proposed techniques to compress the KV cache to 4-bit or even 2-bit formats. However, these techniques aim to reduce the memory footprint of KV cache, but rely on time-intensive decompression to FP16 before executing the attention mechanism using acceleration mechanisms such as FlashAttention.

Additionally, these methods frequently employ dynamic reordering of outlier channels or smooth attention~\cite{lin2024qserve} techniques to adjust key-query interactions, aiming to mitigate outliers~\cite{dettmers2022llmint8}. However, these strategies introduce additional latency, hindering the adoption of integer inference within the attention mechanism. Specifically, the smoothing factor\citep{smoothquant} that transfers outliers from key to query is not applicable in cases where both tensors are quantized activations. Similarly, dynamic reordering requires synchronized reordering of query, key, and value components, reducing its overall effectiveness. Thus, there is a huge need to develop a quantization technique that will enable a quantized execution of the attention operation in a FlashAttention compatible manner and reduce the memory footprint of KV cache while not losing the accuracy of LLMs.

{\it{\ouralg}} thus presents a novel unified technique for enabling quantized execution of attention along with a cooperative KV cache compression mechanism which reduces latency, memory footprint, incurs negligible accuracy loss. We present the in two parts; {\it{\ourquant}} (\autoref{sec:flashq}), which enables quantization of Key, Value, and Query in a FlashAttention compatible manner and {\it{\oursoftmax}}, which enables efficient GPU tensor core-friendly, softmax computation (\autoref{sec:sas}).

%% file: Section/FlashQuant.tex
\section{FlashQ: Headwise mix-precision progressive quantization}
\label{sec:flashq}





{\it{\ourquant}} involves three key components that enable and significantly accelerate the quantized attention mechanism:
\begin{enumerate}
\setlength{\itemsep}{0pt}     
\vspace{-0.9\baselineskip}       
    \item \textbf{Blockwise Progressive Quantization.} FlashAttention-compatible quantization of Key and Value for quantized execution of MatMuls.
    \item \textbf{Head-wise Mixed Precision.} Compress different heads to different formats (2-bit and 4-bit) for maximum compression.
    \item \textbf{Enhanced KV cache Buffer.} For dynamic management of the quantized KV cache during the decode phase, eliminating the need for recompression of KV cache.
\vspace{-0.9\baselineskip}
\end{enumerate}
\autoref{fig:workflow} illustrates the overall flow of {\it{\ourquant}} along with {\it{\oursoftmax}} (discussed later in \autoref{sec:sas}), integrated with the flash attention mechanism.

\subsection{Blockwise Progressive Quantization (BPQ)}
As we saw in \autoref{sec:attn_mechanism}, multi-head attention generates large intermediate activations, which place a significant demand on memory bandwidth. FlashAttention \citep{dao2022flashatten} addresses this by chunking the Query, Key, and Value matrices along the token dimension and tiling the attention mechanism to accelerate the process. FlashAttention divides the $h$-th head's query, key, and value matrices ($\mQh$, $\mKh$, and $\mVh$) into sub-blocks of size $T_c$  and $T_r$, denoted as $\mQh_c$, $\mKh_r$, and $\mVh_r$, for efficient tiled computation. 

To enable a FlashAttention-compatible quantized execution of attention we propose Blockwise Progressive Quantization (BPQ). In BPQ, each of the sub-block is first compressed using 8-bit symmetric quantization. Unlike Qserve, which applies per-channel PQ to weights, we apply BPQ at a sub-block granularity via the function:
\vspace*{-3mm}
\begin{equation}
X^{q1} = Quant^8_{sym}(X)
\label{eq:step1bpq}
\end{equation}
\vspace*{-5mm}

where $\mathbf{X} \in {\mQh_c, \mKh_r, \mVh_r}$ represents a sub-block, and $\text{Quant}_{\text{sym}}^8$ denotes symmetric 8-bit quantization. Given the critical memory bottleneck posed by the KV cache, as described in \autoref{sec:motivation}, we apply progressive quantization after computation to further compress the Key and Value tensors for storage.

\begin{figure}[t!]  
    \centering
        \centering
        \includegraphics[width=0.48\textwidth]{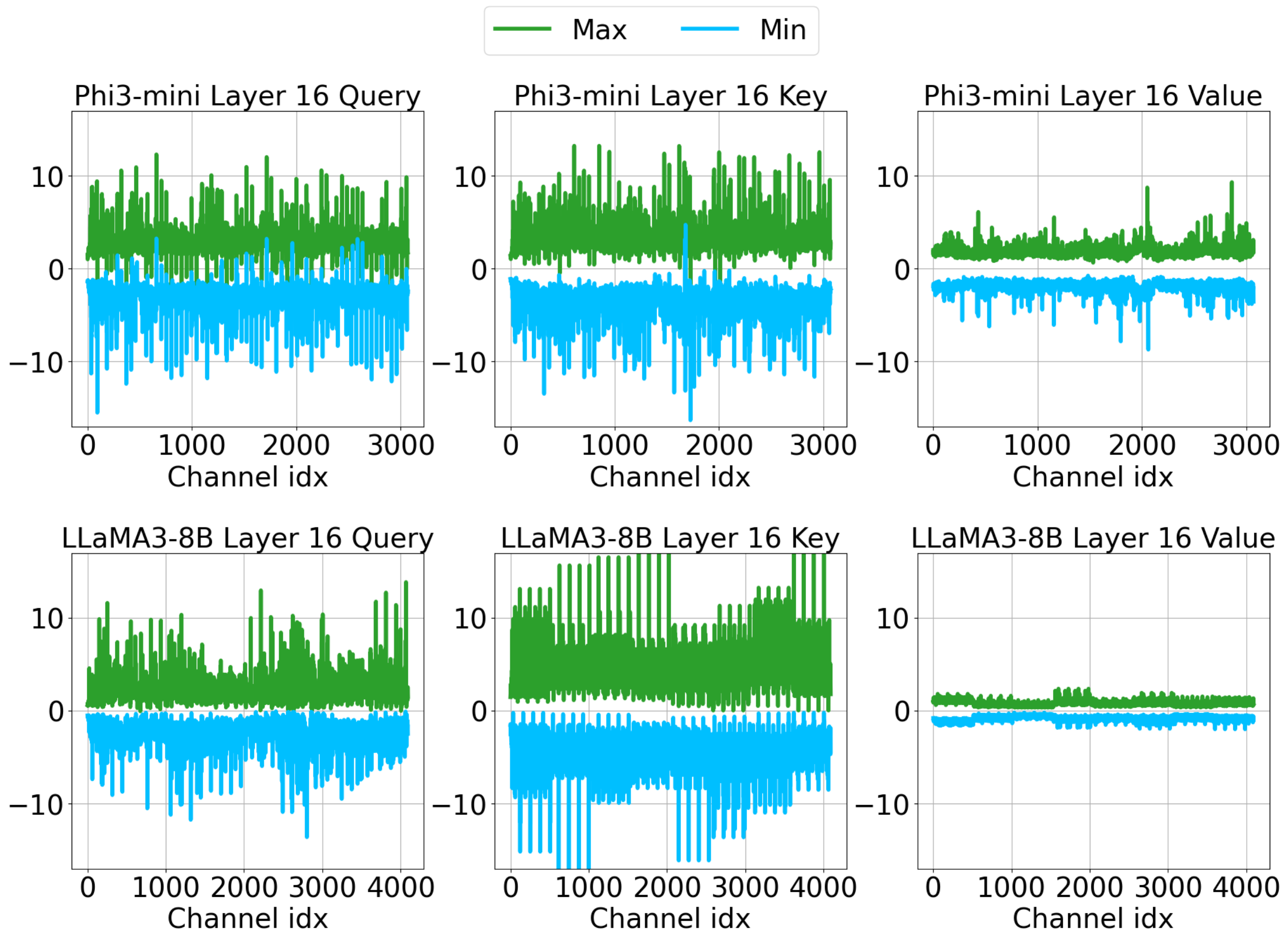}  
    \caption{Query, key and value channels min-max distribution of Phi3-mini and LLaMA3-8B Models. We observe that certain heads in query and key have a number of large-magnitude channels. For value, there is no obvious outlier pattern.
    }
    \label{fig:qkv_distribution}
    \vspace{-6mm}
\end{figure}

Further, building on the techniques from KIVI \citep{liu2024kivi} and informed by our own analysis (\autoref{fig:qkv_distribution}), we further reduce quantization errors by compressing the Key and Value 8-bit tensors in a channel-wise manner using asymmetric quantization:
\begin{equation}
K^{q2}_{g} = Quant^{\text{4 / 2}}_{asym}(K^{q1}_{g}),
V^{q2}_{g} = Quant^{\text{4 / 2}}_{asym}(V^{q1}_{g})
\label{eq:step2bpq}
\end{equation}
where $\hat{K}^{q1}_{g}$ and $\hat{V}^{q1}_{g}$ are groups in each channel of the key and value 8-bit tensors. This minimizes the errors caused by outliers in certain channels while keeping the tiled nature needed to make it compatible with FlashAttention's core algorithm. Since a new query vector is generated at each decoding step, compressing it further beyond the initial 8-bit quantization is unnecessary.

\subsection{Head-wise Mixed Precision}
\label{sec:hwp}
To further enhance progressive quantization and achieve substantial memory savings in the KV cache, we explore reducing the bit-width to 2-bit. Although 4-bit KV cache quantization has shown near-lossless performance across various models, as demonstrated by Atom \citep{zhao2024atom}, QuaRot \citep{ashkboos2024quarot}, and Qserve \citep{lin2024qserve}, uniformly applying 2-bit compression to all attention heads often leads to significant model performance degradation. Approaches like smooth factors \citep{lin2024qserve, smoothQ} and offline reordering (e.g., Qserve and Atom) can improve compression accuracy but introduce additional latency overhead, making them challenging to integrate with dynamically expanding KV caches. While token-wise mixed precision could further enhance accuracy, it incurs dynamic execution overhead in CUDA kernels and is not FlashAttention-compatible, thus severely reducing hardware efficiency.

Inspired by recent work on head pruning in multi-head attention \citep{ge2024modeltell, liu2023dejavu, rajput2024mixattention}, we propose a headwise mixed-precision approach. This approach combines 2-bit and 4-bit compression, and thus achieves a favorable trade-off between accuracy and memory savings. It applies 2-bit and 4-bit BPQ for different heads based on their priority. The priority is calculated by:

\vspace*{-5mm}
\begin{equation}
\text{priority}^{(h)} = \text{gap}^{(h)} \times \text{std}^{(h)}
\end{equation}
\vspace*{-8mm}

Here, $\text{gap}^{(h)}$ represents the difference between the maximum and minimum values across all channels for head $h$, which captures the range of values in the attention mechanism. A larger gap indicates that compressing this head may introduce a larger quantization error, thus making it more important to preserve precision. $\text{std}^{(h)}$ is the standard deviation of the channel-wise gaps within each head. This measures the variability of the gaps, where higher variability (i.e., a larger standard deviation) implies that the head’s distribution is more uneven. Heads with a higher $\text{std}^{(h)}$ are more sensitive to quantization, and thus require higher bit precision to minimize performance degradation.

Using this metric, we rank all heads in terms of their priority scores. Instead of using a fixed threshold, we compress the lowest-priority $\nhu$ heads in each layer to 2-bit, while the remaining heads are quantized to 4-bit. The quantization strategy is thus defined as:

\vspace*{-8mm}
\begin{equation}
\text{Quantization strategy} =
\begin{cases}
\text{2-bit,} & \text{if rank(priority}^{(h)}) \leq \nhu \\
\text{4-bit,} & \text{if rank(priority}^{(h)}) > \nhu \\
\end{cases}
\end{equation}

Here, $\nhu$ is the number of heads to be compressed to 2-bit, determined per layer. By ranking the heads based on their priority and applying lower-bit quantization to the least critical heads (i.e., those with the smallest priority), we achieve a balanced trade-off between memory savings and model performance. \autoref{fig:workflow} illustrates the head-wise quantization dataflow of {\it{\ourquant}, where the second head is block-wise progressively quantized to INT2}. As we will see later, this technique presents a robust way to compress the KV-cache and can be employed in a majority of models easily.




\subsection{Enhanced KV cache Buffer}
\label{sec:decode_buffer}

{\it{\ourquant}} introduces an efficient decoding buffer to accelerate inference during long-context generation. To support integer inference in the attention mechanism, newly generated tokens are temporarily stored in a buffer $ \Bcal $ of size $\nbu$ (e.g.,~$\nbu=64$) and quantized using 8-bit symmetric quantization (\autoref{eq:step1bpq}). This avoids the need for compression at every decoding iteration. Once the buffer reaches its capacity, {\it{\ourquant}} applies progressive quantization (\autoref{eq:step2bpq}) to further compress the keys and values from 8 bits to lower bit-widths for improved memory efficiency.

To ensure stability, we employ a universal scale for 8-bit quantization and clamp outliers that exceed this range, preventing the need to re-compress previously stored tokens in the buffer when new tokens with larger values are added. This approach contrasts with previous methods like KIVI \citep{liu2024kivi} and GEAR \citep{kang2024gear}, which keep buffers in full precision, introducing significant memory overhead and preventing the use of integer inference in the attention mechanism. We summarize the detailed algorithm of {\ouralg} in  Appendix~\ref{app:gear_algorithm}.

%% file: Section/SAS.tex
\section{SAS: Sparse Activated Softmax}
\label{sec:sas}
Softmax is a key bottleneck in the attention mechanism, particularly in flash attention workflows, where tiling operations introduce additional overhead. From our experiments, we observe that softmax computation costs over 30\% of the attention execution time both in prefill and decoding stages. The primary performance issue stems from frequent data type conversions between FP16 and FP32 for performing the exponentiation operation. Current GPUs are limited to single-precision floating-point (FP32) operations for exponentiation, which exacerbates this bottleneck. Previous methods, such as \citep{vasyltsov2021} and \citep{NIPS2017_4e2a6330}, either introduce significant errors that degrade performance or are unsuitable for modern large language models.

To mitigate these issues, we propose a softmax approximation technique based on polynomial approximation(POLY) and lookup tables(LUT), called Sparse Activated Softmax (SAS). This method divides the exponential computation into two parts: the integer and decimal components. For example, an exponent computation can be separated into integer $x_{int}$ and decimal $x_{dec}$ parts which lie in [0,1) (Detailed algorithm in \autoref{appendix:sas}). For the integer part, we use a lookup table, which remains compact because large values in the exponential function decay quickly, allowing us to skip larger integers. For the decimal part, we employ a simple polynomial approximation. 
\begin{equation}
\small e^{-x} = e^{-x_{int}} \times e^{-x_{dec}} \approx LUT(-x_{int}) \times POLY(-x_{dec})
\end{equation}
As such, the approximation algorithm of SAS goes:
\begin{equation}
SAS(x) =
\begin{cases}
    \displaystyle  0, & \text{$x < n_r$} \\
    \displaystyle  LUT(x_{int}) \times POLY(x_{dec}), & \text{$x \geq n_r$}
\end{cases}
\end{equation}


\begin{figure}[t]

\begin{center}
\centerline{\includegraphics[width=0.75\columnwidth]{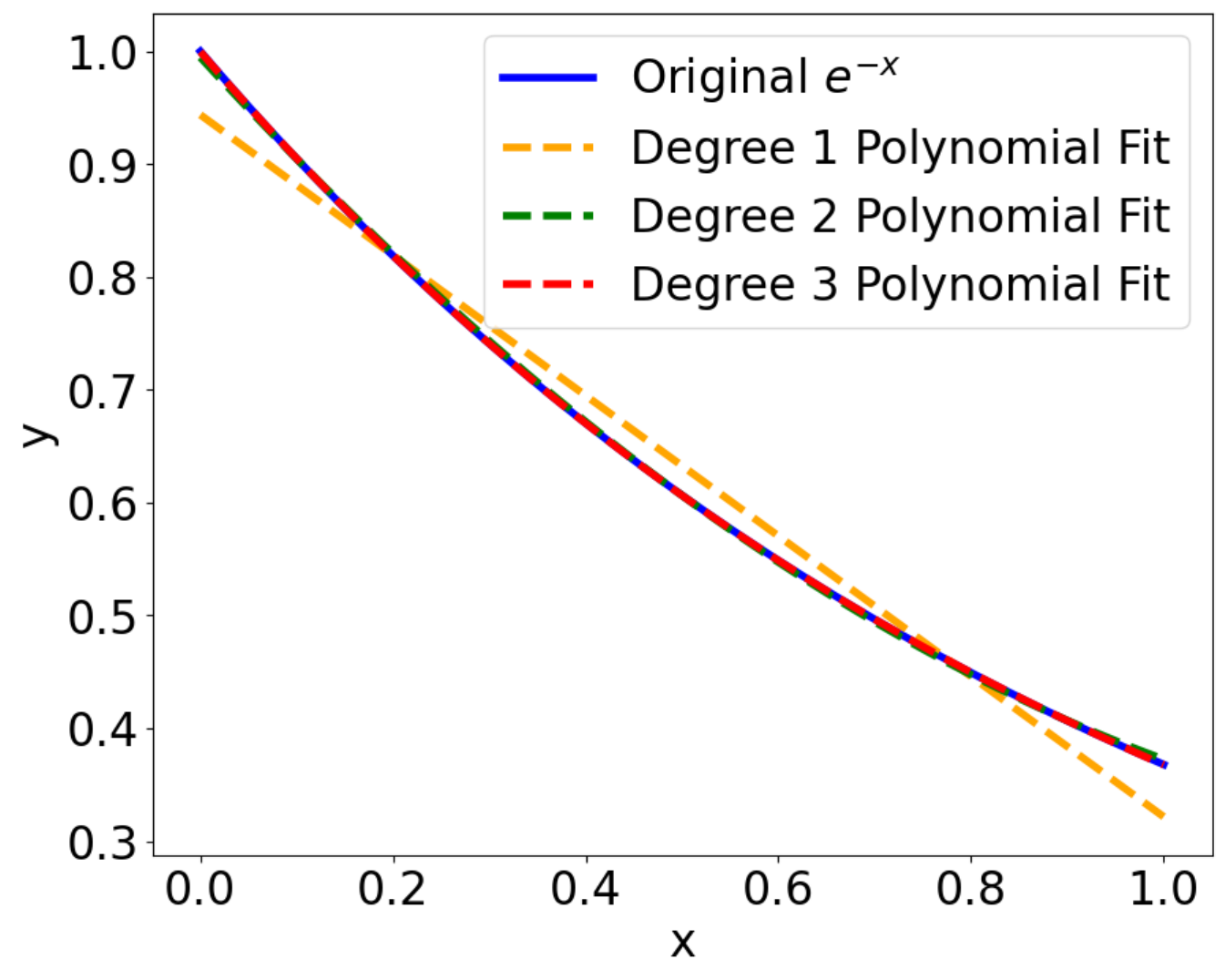}}
\vspace{-4mm}
\caption{Polynomial-fit for the decimal part of value in exponentiation operation.}
\vspace{-10mm}
\label{fig:exponentapprox}
\end{center}

\end{figure}

To optimize the computation of the exponential function $e^{-x}$, we limit the approximation range to \([0, 1]\) and reduce the polynomial degree to less than 3 to enhance computational efficiency. Using the least squares method to determine the coefficients, our 3-degree polynomial approximation of $e^{-x}$ within this range is as follows:
\begin{equation}
\small \displaystyle POLY(x) = -0.1025 \, x^3 + 0.4626 \, x^2 - 0.9922 \, x + 0.9996
\end{equation}
This polynomial captures the essential behavior of \( e^{-x} \) over \([0, 1]\) with minimal computational overhead, as shown in \autoref{fig:exponentapprox}. This hybrid approach reduces computational and data type conversion overhead while preserving accuracy, effectively alleviating the exponentiation bottleneck in flash attention algorithm.
The overall exponentiation operation is accelerated, as the polynomial operation can be done solely in GPU Tensor Cores in FP16 dataformat.

Further, from our observations, the large negative values of the attention scores (resulting from the query-key multiplication) become extremely small after softmax due to the nature of the exponential function. As such, we apply a sparsification strategy, retaining only the larger attention scores within a certain range $n_r$(e.g., 0 to -5) and setting the rest to zero after the sparse activated softmax computation to keep the look-up table small enough. This further reduces the computational cost, making SAS a practical and efficient solution for large-scale language models.

%% file: Section/Evaluations.tex
\section{Evaluations}
\label{sec:evaluations}

{
\setlength{\tabcolsep}{0.3em}
\renewcommand{\arraystretch}{1.2}
\begin{table*}[t!]
\begin{center}
\begin{small}
\begin{tabular}{l|c|ccc|ccc|ccc|c}
\toprule
\multicolumn{2}{c}{\bf Model} 
& \multicolumn{3}{|c}{\bf LLaMA3-8B-inst} 
& \multicolumn{3}{|c|}{\bf Qwen2-7B-inst}
& \multicolumn{3}{|c}{\bf Phi3-3.8B-inst}
& \multicolumn{1}{|c}{\bf All}
\\
\midrule
\multirow{2}*{\bf Method} 
& \multirow{1}*{\bf Bit} 
& {\bf\footnotesize GSM8k} 
& {\bf\footnotesize AQuA} 
& {\bf\footnotesize BBH}
& {\bf\footnotesize GSM8k} 
& {\bf\footnotesize AQuA} 
& {\bf\footnotesize BBH}
& {\bf\footnotesize GSM8k} 
& {\bf\footnotesize AQuA} 
& {\bf\footnotesize BBH}
& {\bf\footnotesize Ave.}
\\
& {\bf $\bb$}
& {Acc} 
& {Acc}
& {Acc}
& {Acc} 
& {Acc}
& {Acc}
& {Acc} 
& {Acc}
& {Acc}
& {Acc}
\\
\midrule
{FP16} 
& {16}
& {78.24}
& {50.79}
& {58.71}
& {71.87}
& {45.67}
& {50.71}
& {84.53}
& {58.66}
& {57.83}
& {61.89}
\\
\midrule
{\footnotesize KIVI\textsubscript{$\gs=64,\nbu=64$}}
& {4}
& {61.18}
& {46.46}
& {53.20}
& {52.16}
& {45.28}
& {45.33}
& {57.09}
& {54.72}
& {51.24}
& {51.85}
\\
{\footnotesize ${\textrm{GEAR-L}}_{\rr=4}^{(\textrm{KCVT})}$}
& {4}
& {64.94}
& {47.09}
& {57.14}
& {54.46}
& {46.06}
& {46.41}
& {79.86}
& {54.36}
& {53.40}
& {55.97}
\\
\midrule
{\footnotesize \textbf{${\textrm{\ouralg}\textsubscript{$\nbu=64$}}$} }
& {4}
& {\textbf{78.31}} 
& {\textbf{48.03}} 
& {\textbf{58.63}} 
& {\textbf{66.19}} 
& {\textbf{46.06}} 
& {\textbf{46.86}}
& {\textbf{84.00}}
& {\textbf{55.91}}
& {\textbf{58.43}}
& {\textbf{60.27}}
\\

\midrule
\midrule
{\footnotesize KIVI\textsubscript{$\gs=64,\nbu=64$}}
& {3} 
& {59.43}
& {44.88}
& {51.40}
& {52.38}
& {41.73}
& {46.47}
& {55.57}
& {47.24}
& {50.98}
& {50.01} 

\\
{\footnotesize ${\textrm{GEAR-L}}_{\rr=4}^{(\textrm{KIVI})}$}
& {3}
& {62.06}
& {46.12}
& {52.72}
& {53.15}
& {\textbf{42.03}} 
& {\textbf{46.65}}
& {56.18}
& {\textbf{49.60}}
& {\textbf{51.39}}
& {51.10}
\\
\midrule 
{\footnotesize ${\textrm{\bf\ouralg}}_{\nbu=64}^{(\textrm{mixed})}$ }
& {2/4}
& {\textbf{77.53}}
& {\textbf{47.94}} 
& {\textbf{54.36}} 
& {\textbf{66.56}}
& {41.91}
& {45.13}
& {\textbf{63.53}}
& {31.49}
& {51.32}
& {\textbf{53.31}}
\\

\bottomrule
\end{tabular}
\end{small}
\end{center}
\vspace{-3mm}
\caption{Results on CoT reasoning tasks, which are hard generative tasks. Here, Bit represents the average compressed bit of KV cache for different methods. The best results are shown in {\bf bold}. Mixed precision results have same KV cache size with 3-bit simulated benchmark.}
\label{tab:maincotresult}
\vspace{-6mm}
\end{table*}
}

\subsection{Experiment Setup}
\textbf{Models and dataset}
We conducted an extensive evaluation of {\it{\ouralg}} for generative inference using a variety of open-source pre-trained and instruction-tuned models, including LLaMA3-8B-inst \citep{dubey2024llama3herdmodels}, Phi3-mini-inst \citep{abdin2024phi3technicalreporthighly}, and Qwen2.5-7B \citep{yang2024qwen2technicalreport}, on multiple generative tasks. These tasks included mathematical reasoning datasets such as GSM8k \citep{cobbe2021training} and AQuA \citep{ling2017program}, symbolic reasoning tasks from BigBench Hard (BBH) \citep{suzgun2022challenging}. Given the complexity of these tasks, we use the chain-of-thought prompts created by \citep{fu2023chainofthought} to improve reasoning, which contains 8-shot demonstrations of multi-step reasoning. 
With the CoT demonstrations, we have the average prefill length of GSM8k, AQuA, and BBH as  900, 1304, {1021} respectively. We then prompt the model to generate 256 tokens and extract answers from them.

\subsection{TurboAttention Implementation}
To minimize overheads, we implemented and optimized {\it{\ouralg}} with OpenAI Triton\citep{Tillet2019TritonAI} kernel support as follows. First, to address the memory bottlenecks caused by frequent quantization and dequantization, we fused the QKV projection with quantization (\autoref{eq:step1bpq}). Additionally, efficient dequantization from progressive quantization was integrated directly into the attention mechanism to reduce decoding latency (\autoref{eq:step2bpq}).

Second, we implement a KV cache buffer (\autoref{sec:decode_buffer} for the newly generated Key and Value vectors during decode. These Key/Value vectors are further compressed to INT4 and INT2 (depending on the head) every $\nbu$ steps. 
To avoid recompression when newly generated KV tokens exceed the previous cache value range maximum, we clamp outliers to ensure only newly generated tokens are compressed to INT8. This further improves decoding efficiency by reducing unnecessary recompression.

Third, we leveraged the Triton framework to implement INT8 matrix multiplications for the attention mechanism, thereby reducing both memory and computation bottlenecks. Finally, we implemented an efficient polynomial approximation and lookup table for {\it{\oursoftmax}}, significantly enhancing the overall performance of the softmax operation. For {\it{\ouralg}}, we fix block size $B_c$, $B_r$, and $n_b$ to 64, {\it{\oursoftmax}} 
threshold $n_r$ to -6, and set half of the heads' KV cache to 2-bit quantization for head-wise mixed precision. \ul{Further optimization can be achieved by implementing CUDA kernels, as they can  deliver higher efficiency and substantial speedup}.

\textbf{Framework.} For our experiments, we applied {\it{\ouralg}} and the baseline methods to models available on \textit{Huggingface} \cite{wolf2019huggingface}, with our inference framework implemented in \textit{PyTorch} \cite{paszke2019pytorch}. Since our focus is on attention acceleration, all other parts of the model are maintained in FP16 unless otherwise stated. 

\textbf{Precision.} In our experiments, we explore ultra-low precision quantization techniques, reporting performance results for 4-bit, 3-bit, and mixed headwise 2/4-bit quantization of the KV cache. 
Our findings indicate that {\it{\ouralg}} achieves near-lossless performance with 4-bit KV cache compression across various tasks and models. However, consistent with findings in other KV cache compression methods, compressing to 2-bit can led to accuracy degradation. To balance accuracy and memory savings, we applied 2-bit precision to $50\%$ of model heads, achieving a practical trade-off. Consequently, we benchmark {\it{\ouralg}} with mixed precision quantization against other 3-bit baseline methods for a comprehensive comparison in \ref{tab:maincotresult}.

Additional ablation studies examining the impact of the number of 2-bit heads and block size variations are provided in \autoref{sec:head wise study}, while the results for pure 2-bit KV cache quantization are included in the Appendix.

\subsection{Baseline Techniques}
We compare {\it{\ouralg}} against the following state-of-the-art techniques:

\noindent$\bullet$ {\it FP16} A dense model where both activations and weights are represented as 16-bit floating-point numbers. This serves as our base dense model without any attention optimizations.

\noindent$\bullet$ {\it FlashAttention}  An attention algorithm applying FlashAttention \citep{shah2024flashattention3} with FP16 computation (and FP32 for exponentiation). This method is "exact" and does not alter accuracy but improves system efficiency by optimizing the attention mechanism.


\noindent$\bullet$ {\it KIVI} \citep{liu2024kivi} A near-lossless KV cache quantization method that compresses the key cache per-channel and the value cache per-token using fine-grained grouping. It stores residual tokens of length $\nbu$ in full precision, and achieving 4-bit KV cache compression at its best accuracy mode.

\noindent$\bullet$ {\it GEAR}\citep{kang2024gear} A concurrent KV cache quantization method that achieves 2-bit near-lossless compression by leveraging low-rank approximations to reduce quantization errors. Residual tokens of length $\nbu$ are stored in full precision for error compensation. Here we use the efficiency version GEAR-L, that deploy quantiations and low rank approximation together to compress KV cache.

\ul{It is important to note that TurboAttention is not just a KV cache compression algorithm but rather a unified approximation algorithm for the attention mechanism}. However, we compare it with KV cache compression algorithms to highlight its ability to preserve both performance and efficiency in large language models (LLMs).

\subsection{Evaluation Results}

\textbf{Generative performance on reasoning tasks.}
\label{sec:pf_reasoning_task}
\autoref{tab:maincotresult}  demonstrate that {\it{\ouralg}} achieves comparable or superior performance to recent KV cache compression algorithms across different models and the challenging reasoning tasks, in both 4-bit and lower-bit compression settings. Despite further compressing the query and approximating the softmax operation, {\it{\ouralg}} maintains near lossless accuracy, achieving an average accuracy of \textbf{60.27\%} across three models and datasets, which is closely aligned with the FP16 baseline average accuracy of \textbf{61.89\%}. Notably, {\it{\ouralg}} also delivers outstanding 3-bit performance, surpassing baseline methods while providing reduced KV cache size, and enhanced inference efficiency.

\begin{figure*}[t]

\begin{center}
\centerline{\includegraphics[width=1.8\columnwidth]{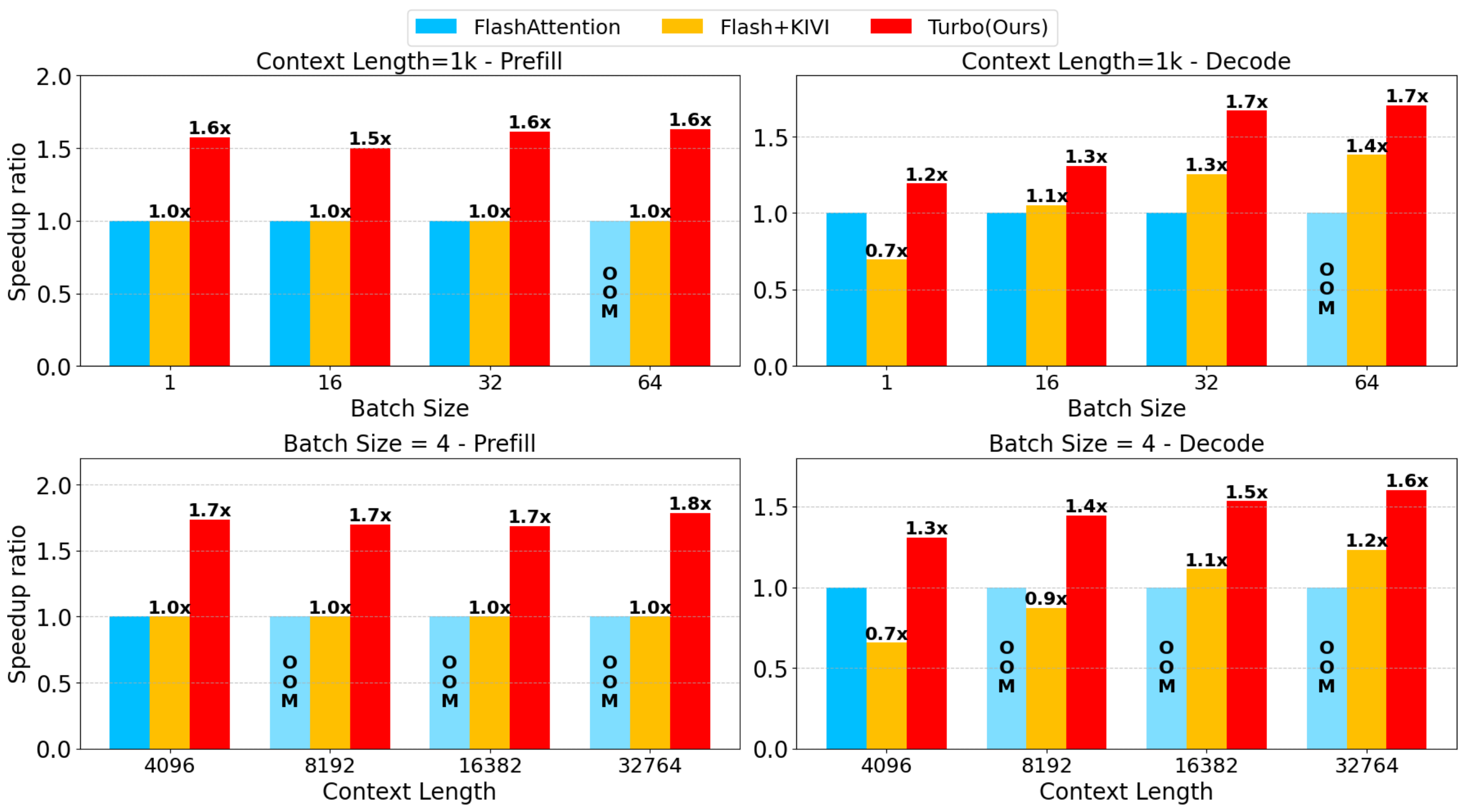}}
\vspace{-4mm}
\caption{Speedup of the attention mechanism of Phi3-medium on 1xA100-80GB-SXM GPU. \textit{\ouralg} achieves up to 1.8$\times$ latency improvement for prefill under various scenarios. For decoding, \textit{\ouralg} delivers up to 1.7$\times$ improvement over FlashAttention in the FP16 data format. Other methods, such as KIVI, may exhibit higher latency than the FP16 baseline due to the time-consuming dequantization process required before computation. OOM denotes out of memory for Phi3-medium model.}
\vspace{-10mm}
\label{fig:system latency profile}
\end{center}

\end{figure*}



\begin{figure}[t!]  
    \centering
    \begin{subfigure}{0.23\textwidth}  
        \centering
        \includegraphics[width=1\textwidth]{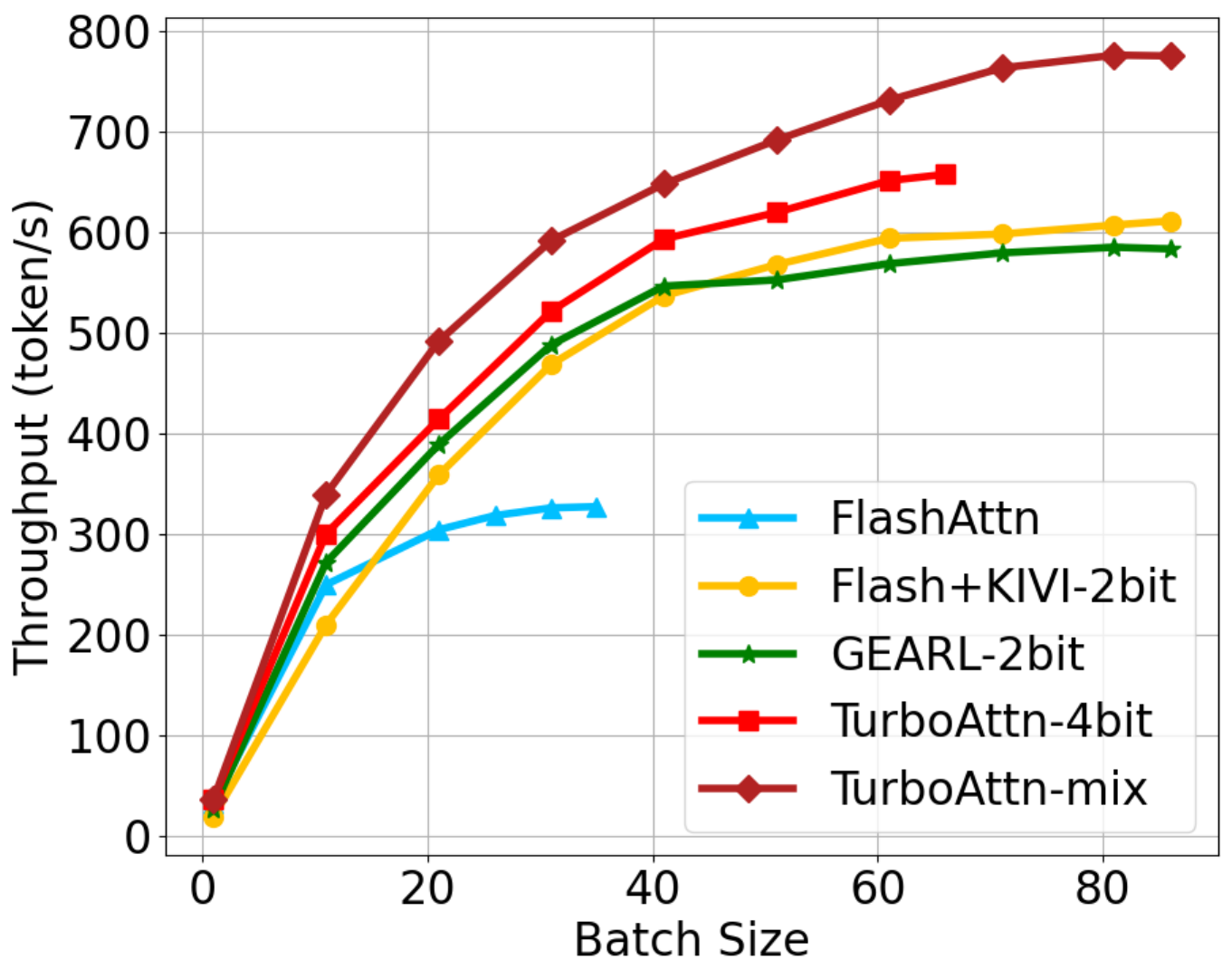}  
        \caption{Delivered throughput}
        \label{fig:throughputs}
    \end{subfigure}
    \hfill  
    \begin{subfigure}{0.23\textwidth}  
        \centering
        \includegraphics[width=1\textwidth]{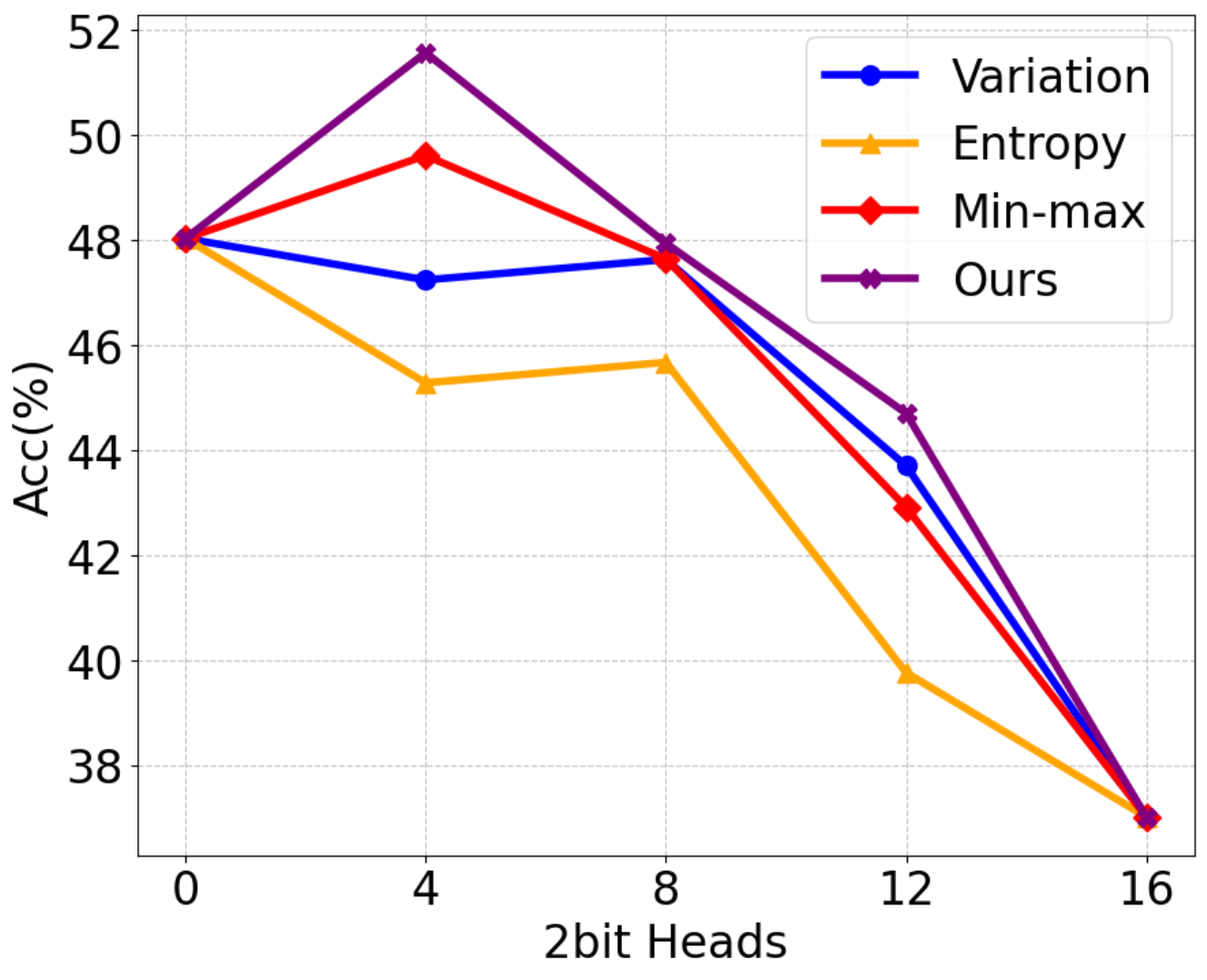}  
        
        \caption{Priority technique vs. Acc.}
        \label{fig:head selection ablation}
    \end{subfigure}
    \caption{(a) Throughput comparison running Phi3-Medium on a 1xA100-SXM-80GB GPU. (b) Comparison of different priority methods with different number of 2-bit heads in LLaMA3-8B-inst on AQua dataset }
    \vspace{-8mm}
    \label{fig:two_subfigures}
\end{figure}

\subsection{Efficiency Comparison}
In this section, we evaluate latency(wall clock time) and maximum throughputs of {\it{\ouralg}} on a single A100 GPU(80GB). 
We measure the attention mechanism of the model under varying batch sizes with a fixed context length of 1k, as well as under different context lengths (ranging from 4k to 32k) with a fixed batch size of 4. We separately assess the prefill and decoding phases and report the speedup ratio compared to the FlashAttention FP16/32 baseline. \autoref{fig:system latency profile} shows the latency comparison across various input settings and methods, all using the same attention mechanism configuration.

Our results indicate that, across batch sizes from 1 to 64 and context lengths from 4k to 32k, {\it{\ouralg}} consistently outperforms both FlashAttention FP16 and other KV cache compression techniques combined with Flash Attention baselines. Specifically, {\it{\ouralg}} achieves a speedup of up to \textbf{1.8×} for prefill and up to \textbf{1.7×} for decoding, highlighting significant improvements in softmax, matrix multiplication, and dequantization efficiency. Additionally, by compressing the KV cache to int-4/2, {\it{\ouralg}} enables long-context generation up to 32k with a batch size of 4, whereas the FP16 baseline encounters out-of-memory (OOM) issues beyond a 4k context length. We also observe that due to the dequantization overhead in other KV cache compression methods, such as KIVI, these methods exhibit worse latency relative to the FP16 baseline.

\autoref{fig:throughputs} shows the throughputs comparison across various batch sizes with fixed context length 1k and generation length 125 tokens. It demonstrats that, compared to the Flash-FP16 baseline, {\it{\ouralg}} significantly improves maximum throughputs by up to \textbf{$2.37\times$}. Meanwhile, {\it{\ouralg}} achieves better throughputs compared to KIVI or GEAR due to our efficient progressive dequantization, compressed KV cache, and {\it{\oursoftmax}} which improves the latency.

\subsection{Ablation Study}
\label{sec:ablation_study}

\textbf{Ablation study on head wise selection methodology}
In this study, we aim to validate the effectiveness of our priority-based head selection strategy as detailed in \autoref{sec:hwp}. We compare our approach against several baseline methods that use simpler, less adaptive metrics to drive head selection. As explored in \autoref{sec:hwp} and illustrated in \autoref{fig:qkv_distribution}, channelwise outliers critically impact quantization error. To address this, we examine head selection strategies that incorporate channelwise value ranges as the basis for selection.

Baseline methods include:
Entropy: Selecting 2-bit heads based on the entropy of each head.
Min-Max: Using the range between minimum and maximum values within heads.
Variation: Evaluating variation in channelwise value ranges.
The benchmarks are run on the LLaMA-3-8B-inst model with 8-shot COT prompts on the AQua dataset. Given LLaMA-3-8B-inst's 8-head configuration per key/value cache, we perform head selection across a range of 0 to 16 heads. Results, as shown in \autoref{fig:head selection ablation}, indicate that our prioritized strategy significantly surpasses other selection approaches in minimizing quantization error, supporting its efficiency and robustness in managing outliers.

\textbf{Ablation study on block size}
\label{sec:head wise study}
We evaluate the sensitivity of {\it{\ouralg}} to varying block sizes ($B_c$ and $B_r$) for the query, key, and value matrices. These block sizes are closely related to the device’s SRAM capacity and have a substantial impact on system efficiency. Consequently, assessing {\it{\ouralg}}'s performance across different block sizes is essential to confirm its robustness and adaptability across configurations. As shown in \autoref{tab:blocksize_ablation}
, we tested Phi3-mini with various block sizes on the GSM8K-COT dataset using an 8-shot setup. The results demonstrate that {\it{\ouralg}} maintains robustness across different block sizes, underscoring its ability to integrate seamlessly with Flash Attention across various systems while preserving model accuracy.


\begin{table}[t!]
\vspace{-0.2mm}
\centering
\caption{{\it{\ouralg}} on Phi3-mini with different block size of query, key, and value.}\label{tab:blocksize_ablation}
\begin{tabular}{llr}
\toprule
              Block size($B_r$,$B_c$) &  Dataset &  Acc \\
\midrule
                (32,32) & GSM8K & 78.01 \\
       (32,64) & GSM8K & 78.01 \\
(64,32) & GSM8K & 78.31 \\
  (64,64) & GSM8K & 78.31 \\
       (64,128) & GSM8K & 78.16 \\
(128,64) & GSM8K & 78.01 \\
  (128,128) & GSM8K & 77.83 \\

\bottomrule

\end{tabular}
\vspace{-4mm}
\end{table}

%% file: Section/Conclusions.tex
\section{Conclusions}
This paper introduces {\it{\ouralg}}, a highly efficient compression algorithm for attention mechanisms that integrates seamlessly with attention acceleration methods like FlashAttention, achieving near loss-less performance. {\it{\ouralg}} demonstrates state-of-the-art results in handling complex generative tasks, including mathematical and reasoning challenges, using 4-bit and 2-bit KV cache compression. Additionally, it provides significant performance gains, with up to 1.8x latency speedup and a 2.37x increase in maximum throughput over the FP16 baseline, marking a substantial advancement in efficient attention mechanism execution.

%% file: Section/Appendix.tex
\onecolumn
\newpage
\section{Detailed Algorithm of {\it{\ouralg}}}
\label{app:gear_algorithm}
\begin{algorithm}[H]
  \caption{\it{\ouralg} Prefill}
  \begin{algorithmic}
    \REQUIRE Matrices $\vQ, \vK, \vV \in \mathbb{R}^{N \times d}$ in HBM, compressed KV cache $\vK^{q2}, \vV^{q2}$,integer scale of KV cache $s_{\vK^{q2}}^{int}$, 
    floating scale $s_{K}$, $s_{V}$,
    $s_{\vV^{q2}}^{int}$,zero point of KV cache $z_{\vK^{q2}}^{int}$, $z_{\vV^{q2}}^{int}$, block sizes $B_c$, $B_r$, Softmax approximate algorithm {\it{\oursoftmax}}.
    
    \STATE \label{alg:stream_attn_split_qkv} Divide $\vQ$ into $T_r = \left\lceil\frac{N}{B_r} \right\rceil$ blocks $\vQ_1, \dots, \vQ_{T_r}$ of size $B_r \times d$ each,
    and divide $\vK, \vV$ in to $T_c = \left\lceil \frac{N}{B_c} \right\rceil$ blocks $\vK_1, \dots, \vK_{T_c}$ and
    $\vV_1, \dots, \vV_{T_c}$, of size $B_c \times d$ each.
    \STATE Divide the output $\vO \in \mathbb{R}^{N \times d}$ into $T_r$ blocks $\vO_i, \dots, \vO_{T_r}$ of size
    $B_r \times d$ each, and divide the logsumexp $L$ into $T_r$ blocks $L_i, \dots, L_{T_r}$ of size
    $B_r$ each.
    \FOR{$1 \le i \le T_r$} \label{alg:stream_attn_outer_loop}
      \STATE \label{alg:stream_attn_load_q} Load $\vQ_i$ from HBM to on-chip SRAM.
      \STATE \label{alg:stream_attn_init} On chip, initialize $\vO_{i}^{(0)} = (0)_{B_r \times d} \in \mathbb{R}^{B_r \times d}, \ell_{i}^{(0)} = (0)_{B_r} \in \mathbb{R}^{B_r}, m_{i}^{(0)} = (-\infty)_{B_r} \in \mathbb{R}^{B_r}$.
      \FOR{$1 \le j \le T_c$}
        \STATE \label{alg:stream_attn_load_kv} Load $\vK_j, \vV_j$ from HBM to on-chip SRAM.
        \STATE On chip, compute 
        $s_{\vQ_i} = \frac{max(abs(\vQ_i))}{119},\vQ_i^{q1} = \left \lceil \frac{\vQ_i}{s_{\vQ_i}} \right \rfloor$,$s_{\vK_j} = \frac{max(abs(\vK_j))}{119},\vK_j^{q1} = \left \lceil \frac{\vK_j}{s_{\vK_j}} \right \rfloor
        ,$\\
        $s_{\vV_j} = \frac{max(abs(\vV_j))}{119},\vV_j^{q1} = \left \lceil \frac{\vV_j}{s_{\vV_j}} \right \rfloor$.
        
        \STATE \label{alg:stream_attn_qk} On chip, compute 
        $\vS_{i}^{(j)} = s_{\vQ_i} \cdot s_{\vK_j} \cdot \vQ_i^{q1} \vK_j^{q1} \in \mathbb{R}^{B_r \times B_c}$.

        \STATE \label{alg:stream_attn_statistics} On chip, compute
        $m_{i}^{(j)} = \mathrm{max}(m_{i}^{(j-1)}, \mathrm{rowmax}(\vS_{i}^{(j)})) \in \mathbb{R}^{B_r}$, $\tilde{\vP}_{i}^{(j)} = SAS(\vS_{i}^{(j)} - m_{i}^{(j)}) \in \mathbb{R}^{B_r \times B_c}$ (pointwise),
        $\ell_{i}^{(j)} = SAS(m_{i}^{j-1} - m_{i}^{(j)}) \ell_{i}^{(j-1)} + \mathrm{row sum}(\tilde{\vP}_{i}^{(j)}) \in \mathbb{R}^{B_r}$.
        \STATE  On chip, compute
        $s_{\tilde{\vP}_{i}^{(j)}} = \frac{max(abs(\tilde{\vP}_{i}^{(j)})}{119},Q(\tilde{\vP}_{i}^{(j)}) = \left \lceil \frac{\tilde{\vP}_{i}^{(j)}}{s_{\tilde{\vP}_{i}^{(j)}}} \right \rfloor$,
        
        \STATE \label{alg:stream_attn_update} On chip, compute
        $\vO_{i}^{(j)} = \diag(SAS({m_{i}^{(j-1)} - m_{i}^{(j)}}))^{-1} \vO_{i}^{(j-1)} + s_{\tilde{\vP}_{i}^{(j)}} \cdot s_{\vV_j} \cdot Q(\tilde{\vP}_{i}^{(j)}) \vV_j^{q1}$.
        \STATE On chip, compute
        $s_{\vK_j^{q2}}^{int} = \left \lceil \frac {max(\vK_j^{q2}) - min(\vK_j^{q2})}{2^{bit} - 1}\right \rfloor$, $z_{\vK_j^{q2}}^{int} = \left \lfloor \frac {min(\vK_j^{q2})}{s_{\vK_j^{q2}}^{int}}\right \rfloor$,
        $\vK_j^{q2} = \left \lceil \frac{\vK_j^{q1}}{s_{\vK_j^{q2}}^{int}} - z_{\vK_j^{q2}}^{int} \right \rfloor$(channelwise)
        \STATE On chip, compute
        $s_{\vV_j^{q2}}^{int} = \left \lceil \frac {max(\vV_j^{q2}) - min(\vV_j^{q2})}{2^{bit} - 1}\right \rfloor$, $z_{\vV_j^{q2}}^{int} = \left \lfloor \frac {min(\vV_j^{q2})}{s_{\vV_j^{q2}}^{int}}\right \rfloor$,
        $\vV_j^{q2} = \left \lceil \frac{\vV_j^{q1}}{s_{\vV_j^{q2}}^{int}} - z_{\vV_j^{q2}}^{int} \right \rfloor$(channelwise)
        \STATE Write $\vK_j^{q2}$ to HBM as the $j$-th block of $\vK^{q2}$.
        \STATE Write $\vV_j^{q2}$ to HBM as the $j$-th block of $\vV^{q2}$.
        \STATE Write $s_{\vK_j^{q2}}^{int}$,$s_{\vV_j^{q2}}^{int}$  to HBM as the $j$-th block of $s_{\vK^{q2}}^{int}$, $s_{\vV^{q2}}^{int}$.
        \STATE Write $s_{\vK_j}$,$s_{\vV_j}$  to HBM as the $j$-th block of $s_{\vK}$, $s_{\vV}$.
        \STATE Write $z_{\vK_j^{q2}}^{int}$,$z_{\vV_j^{q2}}^{int}$  to HBM as the $j$-th block of $z_{\vK^{q2}}^{int}$, $z_{\vV^{q2}}^{int}$.
      \ENDFOR
      \STATE On chip, compute $\vO_{i} = \diag(\ell_{i}^{(T_c)})^{-1} \vO_{i}^{(T_c)}$.
      \STATE On chip, compute $L_{i} = m_{i}^{(T_c)} + \log(\ell_i^{(T_c)})$.
      \STATE Write $\vO_{i}$ to HBM as the $i$-th block of $\vO$.
      \STATE Write $L_{i}$ to HBM as the $i$-th block of $L$.
    \ENDFOR
    \STATE Return the output $\vO$ and the logsumexp $L$.
  \end{algorithmic}
\end{algorithm}
\vspace{-0.5em}

\begin{algorithm}[H]
  \caption{\it{\ouralg} Decode}
  \begin{algorithmic}
    \REQUIRE Matrices $\vQ, \in \mathbb{R}^{1 \times d}$ in HBM, compressed KV cache $\vK^{q2}, \vV^{q2} \in \mathbb{R}^{N \times d}$, integer scale of KV cache $s_{\vK^{q2}}^{int}$,$s_{\vV^{q2}}^{int}$, floating scale $s_{K}$, $s_{V}$, zero point of KV cache $z_{\vK^{q2}}^{int}$, $z_{\vV^{q2}}^{int}$, block size $B_c$, Softmax approximate algorithm {\it{\oursoftmax}}.
    
    \STATE \label{alg:stream_attn_split_qkv} Divide $\vK^{q2}, \vV^{q2}$ in to $T_c = \left\lceil \frac{N}{B_c} \right\rceil$ blocks $\vK_1^{q2}, \dots, \vK_{T_c}^{q2}$ and
    $\vV_1^{q2}, \dots, \vV_{T_c}^{q2}$, of size $B_c \times d$ each.
    
    \STATE Divide $s_{\vK^{q2}}^{int}$,$s_{\vV^{q2}}^{int}$,$z_{\vV^{q2}}^{int}$, block size $B_c$into $T_c = \left\lceil \frac{N}{B_c} \right\rceil$ blocks accordingly .
    
    \STATE Divide the output $\vO \in \mathbb{R}^{1 \times d}$ into $T_c$ blocks $\vO_i, \dots, \vO_{T_c}$ of size
    $1 \times \frac{d}{T_c}$ each, and divide the logsumexp $L$ into $T_r$ blocks $L_i, \dots, L_{T_c}$ of size
    $1$ each.
    
      \STATE \label{alg:stream_attn_load_q} Load $\vQ$ from HBM to on-chip SRAM.
      \STATE \label{alg:stream_attn_init} On chip, initialize $\vO_{j} = (0)_{1 \times \frac{d}{T_c}} \in \mathbb{R}^{1 \times d}, \ell_{i}^{(0)} = (0)_{B_c} \in \mathbb{1}^{B_c}, m_{i}^{(0)} = (-\infty)_{B_c} \in \mathbb{R}^{B_c}$.
      \FOR{$1 \le j \le T_c$}
        \STATE \label{alg:stream_attn_load_kv} Load $\vK_j^{q2}, \vV_j^{q2}$ from HBM to on-chip SRAM.
        \STATE On chip, compute 
        $s_{\vQ} = \frac{max(abs(\vQ))}{119},\vQ^{q1} = \left \lceil \frac{\vQ}{s_{\vQ}} \right \rfloor$
        $\vK_j^{q1} = \vK_j^{q2} \cdot s_{\vK^{q2}_j}^{int} + z_{\vK^{q2}_j}$,
        $\vV_j^{q1} = \vV_j^{q2} \cdot s_{\vV^{q2}_j}^{int} + z_{\vV^{q2}_j}$
        
        \STATE \label{alg:stream_attn_qk} On chip, compute 
        $\vS_j = s_{\vQ} \cdot s_{\vK_j} \cdot \vQ^{q1} \vK_j^{q1} \in \mathbb{R}^{1 \times B_c}$.

        \STATE \label{alg:stream_attn_statistics} On chip, compute
        $m_j = \mathrm{max}(m_{(j-1)}, \mathrm{rowmax}(\vS_j)) \in \mathbb{R}^{1}$, $\tilde{\vP}^{(j)} = SAS(\vS_j - m_j) \in \mathbb{R}^{1 \times B_c}$ (pointwise),
        $\ell^{(j)} = SAS(m^{j-1} - m^{(j)}) \ell^{(j-1)} + \mathrm{row sum}(\tilde{\vP}^{(j)}) \in \mathbb{R}^{1}$.
        
        \STATE  On chip, compute
        $s_{\tilde{\vP}^{(j)}} = \frac{max(abs(\tilde{\vP}^{(j)})}{119},Q(\tilde{\vP}^{(j)}) = \left \lceil \frac{\tilde{\vP}^{(j)}}{s_{\tilde{\vP}^{(j)}}} \right \rfloor$,
        
        \STATE \label{alg:stream_attn_update} On chip, compute
        $\vO^{(j)} = \diag(SAS({m^{(j-1)} - m^{(j)}}))^{-1} \vO^{(j-1)} + s_{\tilde{\vP}^{(j)}} \cdot s_{\vV_j} \cdot Q(\tilde{\vP}^{(j)}) \vV_j^{q1}$.
      \ENDFOR
      \STATE On chip, compute $\vO = \diag(\ell^{(T_c)})^{-1} \vO^{(T_c)}$.
      \STATE On chip, compute $L = m^{(T_c)} + \log(\ell^{(T_c)})$.
      \STATE Write $\vO$ to HBM.

    \STATE Return the output $\vO$ and the logsumexp $L$.
  \end{algorithmic}
\end{algorithm}
\vspace{-0.5em}

    




   
\section{Detailed Algorithm of {\it{\oursoftmax}}}
\label{appendix:sas}
\begin{algorithm}[H] 
\caption{ Sparsity-based Softmax Approximation} \begin{algorithmic} 
\REQUIRE Input matrix $X \in \mathbb{R}^{N \times d}$, integer threshold $n_r$, look-up table $T$ with $T[n_r + 1]$ set to $0$, polynomial approximator function $POLY$.
\STATE \textbf{Step 1: Normalize input values}
\STATE $x_{max} = \text{rowmax}(X)$ \hfill \textit{// Max value per row}
\STATE $X = X - x_{max}$

\STATE \textbf{Step 2: Apply sparsity threshold}
\STATE $X[X < n_r] = n_r + 1$

\STATE \textbf{Step 3: Separate integer and decimal parts}
\STATE $X = X_{int} + X_{decimal}$

\STATE \textbf{Step 4: Approximate exponential values}
\STATE $X_{lut} = T[X_{int}]$ \hfill \textit{// Lookup for integer part}
\STATE $X_{approx} = POLY(X_{decimal})$ \hfill \textit{// Polynomial approx. for decimal part}
\STATE $X = X_{lut} \times X_{approx}$

\STATE \textbf{Step 5: Normalize with row-wise sum}
\STATE $X_{sum} = \text{rowsum}(X)$
\STATE $X = \frac{X}{X_{sum}}$

\STATE Return $X$
\end{algorithmic} 
\end{algorithm}

\vspace{-0.5em}


\section{Detailed illustration of {\it{\ourquant}} and {\it{\ouralg}}}
In this section, we separately evaluate the accuracy degradation caused by {\it{\ourquant}} and {\it{\oursoftmax}}. Results are shown in \autoref{tab:SAS+FlashQ}. It demonstrated that both techniques are near loss-less.
\begin{table}[t!]
\centering
\vspace{-2mm}
\caption{Separate discussion about accuracy degradation of FlashQ and SAS}\label{tab:SAS+FlashQ}
\begin{tabular}{llrl}
\toprule
Model & Dataset & Method & Acc \\ 
\midrule
LLaMA3-8B-inst & AQuA & FP16 & 50.79 \\
LLaMA3-8B-inst & AQuA & FlashQ-4bit & 49.60 \\
LLaMA3-8B-inst & AQuA & SAS & 50.12 \\
LLaMA3-8B-inst & AQuA & FlashQ-4bit + SAS & 48.03 \\
\bottomrule
\end{tabular}

\end{table}

\section{Distribution difference of models}
Different classes of models exhibit distinct query, key, and value (QKV) distributions, which can significantly influence the effectiveness of compression algorithms. In \autoref{fig:qkv_distribution}, we observe that the value cache of the Phi-3 model demonstrates a more pronounced outlier distribution compared to LLaMA and other models. Furthermore, among LLaMA-3, Qwen-2, and Phi-3, the Phi-3 value cache exhibits a stronger outlier distribution along the channel dimension than the token dimension.

Previous approaches, such as GEAR and KIVI, often suffer from performance degradation across these models due to their suboptimal value cache quantization strategies, which primarily rely on groupwise or tokenwise quantization. These methods fail to adequately account for the unique distributional characteristics of certain models, such as Phi-3.

\begin{figure*}[t] \begin{center} \centerline{\includegraphics[width=0.99\columnwidth]{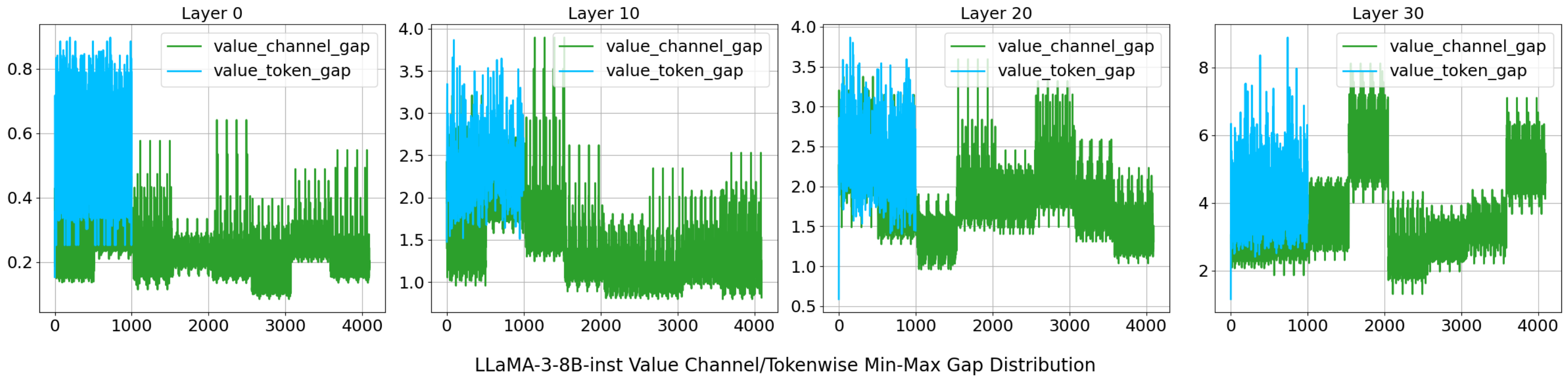}} \vspace{-4mm} \caption{LLaMA-3-8B-inst Value Channel/Tokenwise Min-Max Gap Distribution. This illustrates the distribution of value cache min-max gaps across tokenwise and channelwise dimensions.} \label{fig:value_distribution} \vspace{6mm} 

\centerline{\includegraphics[width=0.99\columnwidth]{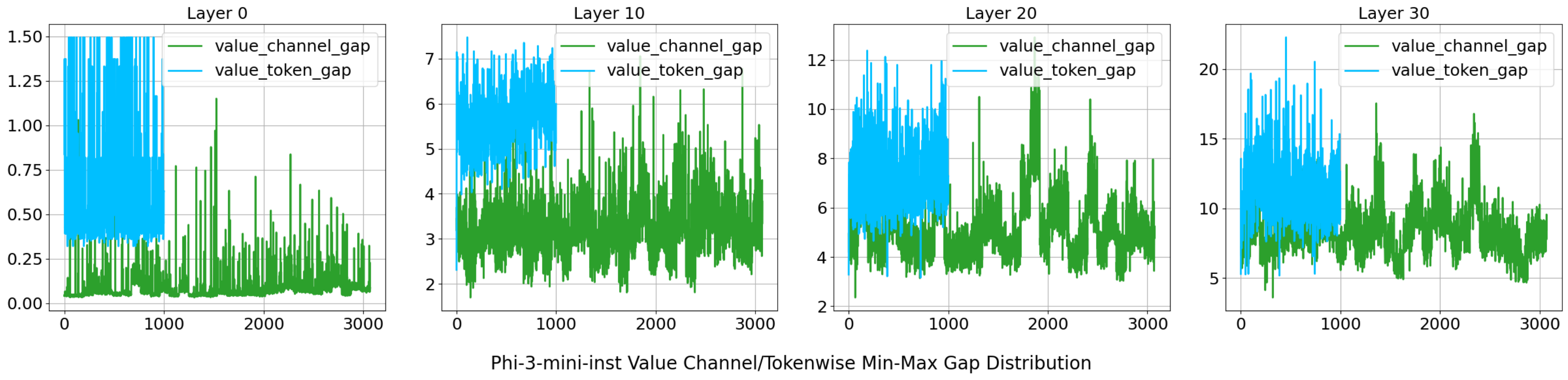}} \vspace{-4mm} \caption{Phi-3-mini-inst Value Channel/Tokenwise Min-Max Gap Distribution. This figure highlights the more significant outlier distribution in the Phi-3 value cache, especially along the channel dimension.}  \label{fig:value_distribution_repeated}

\vspace{2mm}
\centerline{\includegraphics[width=0.65\columnwidth]{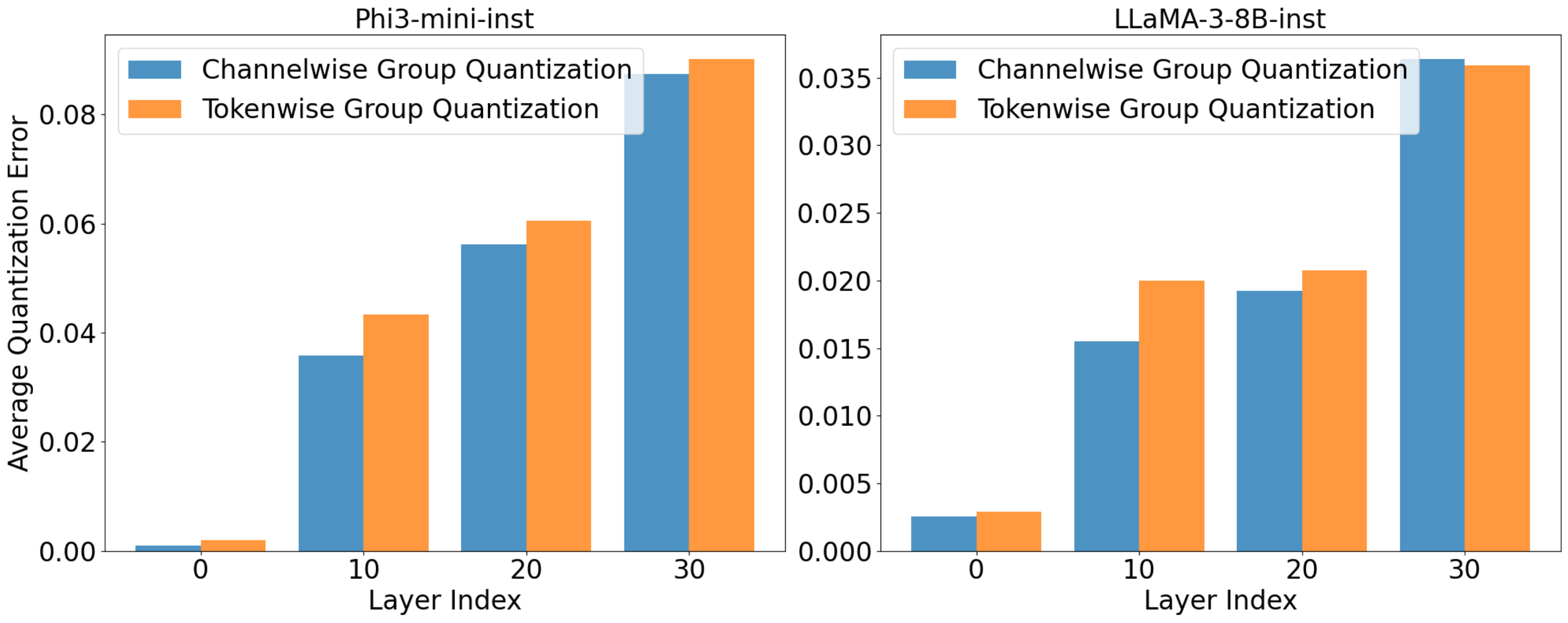}} \vspace{-4mm} \caption{Quantization error of two method}  \label{fig:value_quant_error}
\end{center} \end{figure*}

As shown in \autoref{fig:value_distribution} and \autoref{fig:value_distribution_repeated}, both LLaMA-3-8B-inst and Phi-3-mini-inst exhibit larger outlier distributions along the channel dimension, with Phi-3 being particularly extreme. These findings suggest that tokenwise compression is likely to perform worse than channelwise compression for models like Phi-3, where channel outliers dominate.
\autoref{fig:value_quant_error} shows a more direct result that channelwise group quantization has less quantization error compared with tokenwise group quantization.

\section{Integration of with other compression algorithms}
We further test \ouralg\ with weight and activation quantization algorithms. Combining with these algorithms will exploit further memory and computation savings. Results are shown in \autoref{tab:weightq_combined}.

\begin{table}[t!]
\centering
\vspace{-2mm}
\caption{Separate discussion about accuracy degradation of FlashQ and SAS}\label{tab:weightq_combined}
\begin{tabular}{llrl}
\toprule
Model & Dataset & Method & Acc \\ 
\midrule
LLaMA3-8B-inst & GSM8K & FP16 & 78.24 \\
LLaMA3-8B-inst & GSM8K & LLM.int8() & 77.94 \\
LLaMA3-8B-inst & GSM8K & LLM.int8() + \ouralg & 77.48 \\
LLaMA2-2-7B & PQ & FP16 & 79.05 \\
LLaMA2-2-7B & PQ & Qserve & 78.07 \\
LLaMA2-2-7B & PQ & Qserve + \ouralg & 77.64 \\
\bottomrule
\end{tabular}

\end{table}

%% file: main.bbl
\begin{thebibliography}{41}
\providecommand{\natexlab}[1]{#1}
\providecommand{\url}[1]{\texttt{#1}}
\expandafter\ifx\csname urlstyle\endcsname\relax
  \providecommand{\doi}[1]{doi: #1}\else
  \providecommand{\doi}{doi: \begingroup \urlstyle{rm}\Url}\fi

\bibitem[Abdin et~al.(2024)Abdin, Aneja, Awadalla, Awadallah, Awan, Bach, Bahree, Bakhtiari, Bao, Behl, Benhaim, Bilenko, Bjorck, Bubeck, Cai, Cai, Chaudhary, Chen, Chen, Chen, Chen, Chen, Cheng, Chopra, Dai, Dixon, Eldan, Fragoso, Gao, Gao, Gao, Garg, Giorno, Goswami, Gunasekar, Haider, Hao, Hewett, Hu, Huynh, Iter, Jacobs, Javaheripi, Jin, Karampatziakis, Kauffmann, Khademi, Kim, Kim, Kurilenko, Lee, Lee, Li, Li, Liang, Liden, Lin, Lin, Liu, Liu, Liu, Liu, Liu, Luo, Madan, Mahmoudzadeh, Majercak, Mazzola, Mendes, Mitra, Modi, Nguyen, Norick, Patra, Perez-Becker, Portet, Pryzant, Qin, Radmilac, Ren, de~Rosa, Rosset, Roy, Ruwase, Saarikivi, Saied, Salim, Santacroce, Shah, Shang, Sharma, Shen, Shukla, Song, Tanaka, Tupini, Vaddamanu, Wang, Wang, Wang, Wang, Wang, Wang, Ward, Wen, Witte, Wu, Wu, Wyatt, Xiao, Xu, Xu, Xu, Xue, Yadav, Yang, Yang, Yang, Yang, Yu, Yuan, Zhang, Zhang, Zhang, Zhang, Zhang, Zhang, Zhang, and Zhou]{abdin2024phi3technicalreporthighly}
Abdin, M., Aneja, J., Awadalla, H., Awadallah, A., Awan, A.~A., Bach, N., Bahree, A., Bakhtiari, A., Bao, J., Behl, H., Benhaim, A., Bilenko, M., Bjorck, J., Bubeck, S., Cai, M., Cai, Q., Chaudhary, V., Chen, D., Chen, D., Chen, W., Chen, Y.-C., Chen, Y.-L., Cheng, H., Chopra, P., Dai, X., Dixon, M., Eldan, R., Fragoso, V., Gao, J., Gao, M., Gao, M., Garg, A., Giorno, A.~D., Goswami, A., Gunasekar, S., Haider, E., Hao, J., Hewett, R.~J., Hu, W., Huynh, J., Iter, D., Jacobs, S.~A., Javaheripi, M., Jin, X., Karampatziakis, N., Kauffmann, P., Khademi, M., Kim, D., Kim, Y.~J., Kurilenko, L., Lee, J.~R., Lee, Y.~T., Li, Y., Li, Y., Liang, C., Liden, L., Lin, X., Lin, Z., Liu, C., Liu, L., Liu, M., Liu, W., Liu, X., Luo, C., Madan, P., Mahmoudzadeh, A., Majercak, D., Mazzola, M., Mendes, C. C.~T., Mitra, A., Modi, H., Nguyen, A., Norick, B., Patra, B., Perez-Becker, D., Portet, T., Pryzant, R., Qin, H., Radmilac, M., Ren, L., de~Rosa, G., Rosset, C., Roy, S., Ruwase, O., Saarikivi, O., Saied, A., Salim, A.,
  Santacroce, M., Shah, S., Shang, N., Sharma, H., Shen, Y., Shukla, S., Song, X., Tanaka, M., Tupini, A., Vaddamanu, P., Wang, C., Wang, G., Wang, L., Wang, S., Wang, X., Wang, Y., Ward, R., Wen, W., Witte, P., Wu, H., Wu, X., Wyatt, M., Xiao, B., Xu, C., Xu, J., Xu, W., Xue, J., Yadav, S., Yang, F., Yang, J., Yang, Y., Yang, Z., Yu, D., Yuan, L., Zhang, C., Zhang, C., Zhang, J., Zhang, L.~L., Zhang, Y., Zhang, Y., Zhang, Y., and Zhou, X.
\newblock Phi-3 technical report: A highly capable language model locally on your phone, 2024.
\newblock URL \url{https://arxiv.org/abs/2404.14219}.

\bibitem[Ashkboos et~al.(2024)Ashkboos, Mohtashami, Croci, Li, Jaggi, Alistarh, Hoefler, and Hensman]{ashkboos2024quarot}
Ashkboos, S., Mohtashami, A., Croci, M.~L., Li, B., Jaggi, M., Alistarh, D., Hoefler, T., and Hensman, J.
\newblock Quarot: Outlier-free 4-bit inference in rotated llms, 2024.
\newblock URL \url{https://arxiv.org/abs/2404.00456}.

\bibitem[Brandon et~al.(2023)Brandon, Nrusimha, Qian, Ankner, Jin, Song, and Ragan-Kelley]{brandon2023striped}
Brandon, W., Nrusimha, A., Qian, K., Ankner, Z., Jin, T., Song, Z., and Ragan-Kelley, J.
\newblock Striped attention: Faster ring attention for causal transformers.
\newblock \emph{arXiv preprint arXiv:2311.09431}, 2023.

\bibitem[Brown et~al.(2020)Brown, Mann, Ryder, Subbiah, Kaplan, Dhariwal, Neelakantan, Shyam, Sastry, Askell, Agarwal, Herbert-Voss, Krueger, Henighan, Child, Ramesh, Ziegler, Wu, Winter, Hesse, Chen, Sigler, Litwin, Gray, Chess, Clark, Berner, McCandlish, Radford, Sutskever, and Amodei]{brown2020languagemodelsfewshotlearners}
Brown, T.~B., Mann, B., Ryder, N., Subbiah, M., Kaplan, J., Dhariwal, P., Neelakantan, A., Shyam, P., Sastry, G., Askell, A., Agarwal, S., Herbert-Voss, A., Krueger, G., Henighan, T., Child, R., Ramesh, A., Ziegler, D.~M., Wu, J., Winter, C., Hesse, C., Chen, M., Sigler, E., Litwin, M., Gray, S., Chess, B., Clark, J., Berner, C., McCandlish, S., Radford, A., Sutskever, I., and Amodei, D.
\newblock Language models are few-shot learners, 2020.
\newblock URL \url{https://arxiv.org/abs/2005.14165}.

\bibitem[Cobbe et~al.(2021)Cobbe, Kosaraju, Bavarian, Chen, Jun, Kaiser, Plappert, Tworek, Hilton, Nakano, Hesse, and Schulman]{cobbe2021training}
Cobbe, K., Kosaraju, V., Bavarian, M., Chen, M., Jun, H., Kaiser, L., Plappert, M., Tworek, J., Hilton, J., Nakano, R., Hesse, C., and Schulman, J.
\newblock Training verifiers to solve math word problems, 2021.

\bibitem[Dao(2023)]{dao2023flashattention2}
Dao, T.
\newblock Flashattention-2: Faster attention with better parallelism and work partitioning, 2023.
\newblock URL \url{https://arxiv.org/abs/2307.08691}.

\bibitem[Dao et~al.(2022)Dao, Fu, Ermon, Rudra, and Ré]{dao2022flashatten}
Dao, T., Fu, D.~Y., Ermon, S., Rudra, A., and Ré, C.
\newblock Flashattention: Fast and memory-efficient exact attention with io-awareness, 2022.
\newblock URL \url{https://arxiv.org/abs/2205.14135}.

\bibitem[Dao et~al.(2023)Dao, Haziza, Massa, and Sizov]{flashdecoding}
Dao, T., Haziza, D., Massa, F., and Sizov, G.
\newblock Flashdecoding: {S}tanford {C}{R}{F}{M} --- crfm.stanford.edu.
\newblock \url{https://crfm.stanford.edu/2023/10/12/flashdecoding.html}, 2023.
\newblock [Accessed 22-04-2024].

\bibitem[Dettmers et~al.(2022)Dettmers, Lewis, Belkada, and Zettlemoyer]{dettmers2022llmint8}
Dettmers, T., Lewis, M., Belkada, Y., and Zettlemoyer, L.
\newblock Llm.int8(): 8-bit matrix multiplication for transformers at scale, 2022.
\newblock URL \url{https://arxiv.org/abs/2208.07339}.

\bibitem[Dodge et~al.(2021)Dodge, Sap, Marasović, Agnew, Ilharco, Groeneveld, Mitchell, and Gardner]{dodge2021c4}
Dodge, J., Sap, M., Marasović, A., Agnew, W., Ilharco, G., Groeneveld, D., Mitchell, M., and Gardner, M.
\newblock Documenting large webtext corpora: A case study on the colossal clean crawled corpus, 2021.
\newblock URL \url{https://arxiv.org/abs/2104.08758}.

\bibitem[Dubey et~al.(2024)Dubey, Jauhri, Pandey, Kadian, Al-Dahle, Letman, Mathur, Schelten, Yang, Fan, Goyal, Hartshorn, Yang, Mitra, Sravankumar, Korenev, Hinsvark, Rao, Zhang, Rodriguez, Gregerson, Spataru, Roziere, Biron, Tang, Chern, Caucheteux, Nayak, Bi, Marra, McConnell, Keller, Touret, Wu, Wong, Ferrer, Nikolaidis, Allonsius, Song, Pintz, Livshits, Esiobu, Choudhary, Mahajan, Garcia-Olano, Perino, Hupkes, Lakomkin, AlBadawy, Lobanova, Dinan, Smith, Radenovic, Zhang, Synnaeve, Lee, Anderson, Nail, Mialon, Pang, Cucurell, Nguyen, Korevaar, Xu, Touvron, Zarov, Ibarra, Kloumann, Misra, Evtimov, Copet, Lee, Geffert, Vranes, Park, Mahadeokar, Shah, van~der Linde, Billock, Hong, Lee, Fu, Chi, Huang, Liu, Wang, Yu, Bitton, Spisak, Park, Rocca, Johnstun, Saxe, Jia, Alwala, Upasani, Plawiak, Li, Heafield, Stone, El-Arini, Iyer, Malik, Chiu, Bhalla, Rantala-Yeary, van~der Maaten, Chen, Tan, Jenkins, Martin, Madaan, Malo, Blecher, Landzaat, de~Oliveira, Muzzi, Pasupuleti, Singh, Paluri, Kardas, Oldham, Rita,
  Pavlova, Kambadur, Lewis, Si, Singh, Hassan, Goyal, Torabi, Bashlykov, Bogoychev, Chatterji, Duchenne, Çelebi, Alrassy, Zhang, Li, Vasic, Weng, Bhargava, Dubal, Krishnan, Koura, Xu, He, Dong, Srinivasan, Ganapathy, Calderer, Cabral, Stojnic, Raileanu, Girdhar, Patel, Sauvestre, Polidoro, Sumbaly, Taylor, Silva, Hou, Wang, Hosseini, Chennabasappa, Singh, Bell, Kim, Edunov, Nie, Narang, Raparthy, Shen, Wan, Bhosale, Zhang, Vandenhende, Batra, Whitman, Sootla, Collot, Gururangan, Borodinsky, Herman, Fowler, Sheasha, Georgiou, Scialom, Speckbacher, Mihaylov, Xiao, Karn, Goswami, Gupta, Ramanathan, Kerkez, Gonguet, Do, Vogeti, Petrovic, Chu, Xiong, Fu, Meers, Martinet, Wang, Tan, Xie, Jia, Wang, Goldschlag, Gaur, Babaei, Wen, Song, Zhang, Li, Mao, Coudert, Yan, Chen, Papakipos, Singh, Grattafiori, Jain, Kelsey, Shajnfeld, Gangidi, Victoria, Goldstand, Menon, Sharma, Boesenberg, Vaughan, Baevski, Feinstein, Kallet, Sangani, Yunus, Lupu, Alvarado, Caples, Gu, Ho, Poulton, Ryan, Ramchandani, Franco, Saraf,
  Chowdhury, Gabriel, Bharambe, Eisenman, Yazdan, James, Maurer, Leonhardi, Huang, Loyd, Paola, Paranjape, Liu, Wu, Ni, Hancock, Wasti, Spence, Stojkovic, Gamido, Montalvo, Parker, Burton, Mejia, Wang, Kim, Zhou, Hu, Chu, Cai, Tindal, Feichtenhofer, Civin, Beaty, Kreymer, Li, Wyatt, Adkins, Xu, Testuggine, David, Parikh, Liskovich, Foss, Wang, Le, Holland, Dowling, Jamil, Montgomery, Presani, Hahn, Wood, Brinkman, Arcaute, Dunbar, Smothers, Sun, Kreuk, Tian, Ozgenel, Caggioni, Guzmán, Kanayet, Seide, Florez, Schwarz, Badeer, Swee, Halpern, Thattai, Herman, Sizov, Guangyi, Zhang, Lakshminarayanan, Shojanazeri, Zou, Wang, Zha, Habeeb, Rudolph, Suk, Aspegren, Goldman, Damlaj, Molybog, Tufanov, Veliche, Gat, Weissman, Geboski, Kohli, Asher, Gaya, Marcus, Tang, Chan, Zhen, Reizenstein, Teboul, Zhong, Jin, Yang, Cummings, Carvill, Shepard, McPhie, Torres, Ginsburg, Wang, Wu, U, Saxena, Prasad, Khandelwal, Zand, Matosich, Veeraraghavan, Michelena, Li, Huang, Chawla, Lakhotia, Huang, Chen, Garg, A, Silva, Bell,
  Zhang, Guo, Yu, Moshkovich, Wehrstedt, Khabsa, Avalani, Bhatt, Tsimpoukelli, Mankus, Hasson, Lennie, Reso, Groshev, Naumov, Lathi, Keneally, Seltzer, Valko, Restrepo, Patel, Vyatskov, Samvelyan, Clark, Macey, Wang, Hermoso, Metanat, Rastegari, Bansal, Santhanam, Parks, White, Bawa, Singhal, Egebo, Usunier, Laptev, Dong, Zhang, Cheng, Chernoguz, Hart, Salpekar, Kalinli, Kent, Parekh, Saab, Balaji, Rittner, Bontrager, Roux, Dollar, Zvyagina, Ratanchandani, Yuvraj, Liang, Alao, Rodriguez, Ayub, Murthy, Nayani, Mitra, Li, Hogan, Battey, Wang, Maheswari, Howes, Rinott, Bondu, Datta, Chugh, Hunt, Dhillon, Sidorov, Pan, Verma, Yamamoto, Ramaswamy, Lindsay, Lindsay, Feng, Lin, Zha, Shankar, Zhang, Zhang, Wang, Agarwal, Sajuyigbe, Chintala, Max, Chen, Kehoe, Satterfield, Govindaprasad, Gupta, Cho, Virk, Subramanian, Choudhury, Goldman, Remez, Glaser, Best, Kohler, Robinson, Li, Zhang, Matthews, Chou, Shaked, Vontimitta, Ajayi, Montanez, Mohan, Kumar, Mangla, Albiero, Ionescu, Poenaru, Mihailescu, Ivanov, Li, Wang,
  Jiang, Bouaziz, Constable, Tang, Wang, Wu, Wang, Xia, Wu, Gao, Chen, Hu, Jia, Qi, Li, Zhang, Zhang, Adi, Nam, Yu, Wang, Hao, Qian, He, Rait, DeVito, Rosnbrick, Wen, Yang, and Zhao]{dubey2024llama3herdmodels}
Dubey, A., Jauhri, A., Pandey, A., Kadian, A., Al-Dahle, A., Letman, A., Mathur, A., Schelten, A., Yang, A., Fan, A., Goyal, A., Hartshorn, A., Yang, A., Mitra, A., Sravankumar, A., Korenev, A., Hinsvark, A., Rao, A., Zhang, A., Rodriguez, A., Gregerson, A., Spataru, A., Roziere, B., Biron, B., Tang, B., Chern, B., Caucheteux, C., Nayak, C., Bi, C., Marra, C., McConnell, C., Keller, C., Touret, C., Wu, C., Wong, C., Ferrer, C.~C., Nikolaidis, C., Allonsius, D., Song, D., Pintz, D., Livshits, D., Esiobu, D., Choudhary, D., Mahajan, D., Garcia-Olano, D., Perino, D., Hupkes, D., Lakomkin, E., AlBadawy, E., Lobanova, E., Dinan, E., Smith, E.~M., Radenovic, F., Zhang, F., Synnaeve, G., Lee, G., Anderson, G.~L., Nail, G., Mialon, G., Pang, G., Cucurell, G., Nguyen, H., Korevaar, H., Xu, H., Touvron, H., Zarov, I., Ibarra, I.~A., Kloumann, I., Misra, I., Evtimov, I., Copet, J., Lee, J., Geffert, J., Vranes, J., Park, J., Mahadeokar, J., Shah, J., van~der Linde, J., Billock, J., Hong, J., Lee, J., Fu, J., Chi, J.,
  Huang, J., Liu, J., Wang, J., Yu, J., Bitton, J., Spisak, J., Park, J., Rocca, J., Johnstun, J., Saxe, J., Jia, J., Alwala, K.~V., Upasani, K., Plawiak, K., Li, K., Heafield, K., Stone, K., El-Arini, K., Iyer, K., Malik, K., Chiu, K., Bhalla, K., Rantala-Yeary, L., van~der Maaten, L., Chen, L., Tan, L., Jenkins, L., Martin, L., Madaan, L., Malo, L., Blecher, L., Landzaat, L., de~Oliveira, L., Muzzi, M., Pasupuleti, M., Singh, M., Paluri, M., Kardas, M., Oldham, M., Rita, M., Pavlova, M., Kambadur, M., Lewis, M., Si, M., Singh, M.~K., Hassan, M., Goyal, N., Torabi, N., Bashlykov, N., Bogoychev, N., Chatterji, N., Duchenne, O., Çelebi, O., Alrassy, P., Zhang, P., Li, P., Vasic, P., Weng, P., Bhargava, P., Dubal, P., Krishnan, P., Koura, P.~S., Xu, P., He, Q., Dong, Q., Srinivasan, R., Ganapathy, R., Calderer, R., Cabral, R.~S., Stojnic, R., Raileanu, R., Girdhar, R., Patel, R., Sauvestre, R., Polidoro, R., Sumbaly, R., Taylor, R., Silva, R., Hou, R., Wang, R., Hosseini, S., Chennabasappa, S., Singh, S.,
  Bell, S., Kim, S.~S., Edunov, S., Nie, S., Narang, S., Raparthy, S., Shen, S., Wan, S., Bhosale, S., Zhang, S., Vandenhende, S., Batra, S., Whitman, S., Sootla, S., Collot, S., Gururangan, S., Borodinsky, S., Herman, T., Fowler, T., Sheasha, T., Georgiou, T., Scialom, T., Speckbacher, T., Mihaylov, T., Xiao, T., Karn, U., Goswami, V., Gupta, V., Ramanathan, V., Kerkez, V., Gonguet, V., Do, V., Vogeti, V., Petrovic, V., Chu, W., Xiong, W., Fu, W., Meers, W., Martinet, X., Wang, X., Tan, X.~E., Xie, X., Jia, X., Wang, X., Goldschlag, Y., Gaur, Y., Babaei, Y., Wen, Y., Song, Y., Zhang, Y., Li, Y., Mao, Y., Coudert, Z.~D., Yan, Z., Chen, Z., Papakipos, Z., Singh, A., Grattafiori, A., Jain, A., Kelsey, A., Shajnfeld, A., Gangidi, A., Victoria, A., Goldstand, A., Menon, A., Sharma, A., Boesenberg, A., Vaughan, A., Baevski, A., Feinstein, A., Kallet, A., Sangani, A., Yunus, A., Lupu, A., Alvarado, A., Caples, A., Gu, A., Ho, A., Poulton, A., Ryan, A., Ramchandani, A., Franco, A., Saraf, A., Chowdhury, A., Gabriel,
  A., Bharambe, A., Eisenman, A., Yazdan, A., James, B., Maurer, B., Leonhardi, B., Huang, B., Loyd, B., Paola, B.~D., Paranjape, B., Liu, B., Wu, B., Ni, B., Hancock, B., Wasti, B., Spence, B., Stojkovic, B., Gamido, B., Montalvo, B., Parker, C., Burton, C., Mejia, C., Wang, C., Kim, C., Zhou, C., Hu, C., Chu, C.-H., Cai, C., Tindal, C., Feichtenhofer, C., Civin, D., Beaty, D., Kreymer, D., Li, D., Wyatt, D., Adkins, D., Xu, D., Testuggine, D., David, D., Parikh, D., Liskovich, D., Foss, D., Wang, D., Le, D., Holland, D., Dowling, E., Jamil, E., Montgomery, E., Presani, E., Hahn, E., Wood, E., Brinkman, E., Arcaute, E., Dunbar, E., Smothers, E., Sun, F., Kreuk, F., Tian, F., Ozgenel, F., Caggioni, F., Guzmán, F., Kanayet, F., Seide, F., Florez, G.~M., Schwarz, G., Badeer, G., Swee, G., Halpern, G., Thattai, G., Herman, G., Sizov, G., Guangyi, Zhang, Lakshminarayanan, G., Shojanazeri, H., Zou, H., Wang, H., Zha, H., Habeeb, H., Rudolph, H., Suk, H., Aspegren, H., Goldman, H., Damlaj, I., Molybog, I.,
  Tufanov, I., Veliche, I.-E., Gat, I., Weissman, J., Geboski, J., Kohli, J., Asher, J., Gaya, J.-B., Marcus, J., Tang, J., Chan, J., Zhen, J., Reizenstein, J., Teboul, J., Zhong, J., Jin, J., Yang, J., Cummings, J., Carvill, J., Shepard, J., McPhie, J., Torres, J., Ginsburg, J., Wang, J., Wu, K., U, K.~H., Saxena, K., Prasad, K., Khandelwal, K., Zand, K., Matosich, K., Veeraraghavan, K., Michelena, K., Li, K., Huang, K., Chawla, K., Lakhotia, K., Huang, K., Chen, L., Garg, L., A, L., Silva, L., Bell, L., Zhang, L., Guo, L., Yu, L., Moshkovich, L., Wehrstedt, L., Khabsa, M., Avalani, M., Bhatt, M., Tsimpoukelli, M., Mankus, M., Hasson, M., Lennie, M., Reso, M., Groshev, M., Naumov, M., Lathi, M., Keneally, M., Seltzer, M.~L., Valko, M., Restrepo, M., Patel, M., Vyatskov, M., Samvelyan, M., Clark, M., Macey, M., Wang, M., Hermoso, M.~J., Metanat, M., Rastegari, M., Bansal, M., Santhanam, N., Parks, N., White, N., Bawa, N., Singhal, N., Egebo, N., Usunier, N., Laptev, N.~P., Dong, N., Zhang, N., Cheng, N.,
  Chernoguz, O., Hart, O., Salpekar, O., Kalinli, O., Kent, P., Parekh, P., Saab, P., Balaji, P., Rittner, P., Bontrager, P., Roux, P., Dollar, P., Zvyagina, P., Ratanchandani, P., Yuvraj, P., Liang, Q., Alao, R., Rodriguez, R., Ayub, R., Murthy, R., Nayani, R., Mitra, R., Li, R., Hogan, R., Battey, R., Wang, R., Maheswari, R., Howes, R., Rinott, R., Bondu, S.~J., Datta, S., Chugh, S., Hunt, S., Dhillon, S., Sidorov, S., Pan, S., Verma, S., Yamamoto, S., Ramaswamy, S., Lindsay, S., Lindsay, S., Feng, S., Lin, S., Zha, S.~C., Shankar, S., Zhang, S., Zhang, S., Wang, S., Agarwal, S., Sajuyigbe, S., Chintala, S., Max, S., Chen, S., Kehoe, S., Satterfield, S., Govindaprasad, S., Gupta, S., Cho, S., Virk, S., Subramanian, S., Choudhury, S., Goldman, S., Remez, T., Glaser, T., Best, T., Kohler, T., Robinson, T., Li, T., Zhang, T., Matthews, T., Chou, T., Shaked, T., Vontimitta, V., Ajayi, V., Montanez, V., Mohan, V., Kumar, V.~S., Mangla, V., Albiero, V., Ionescu, V., Poenaru, V., Mihailescu, V.~T., Ivanov, V., Li,
  W., Wang, W., Jiang, W., Bouaziz, W., Constable, W., Tang, X., Wang, X., Wu, X., Wang, X., Xia, X., Wu, X., Gao, X., Chen, Y., Hu, Y., Jia, Y., Qi, Y., Li, Y., Zhang, Y., Zhang, Y., Adi, Y., Nam, Y., Yu, Wang, Hao, Y., Qian, Y., He, Y., Rait, Z., DeVito, Z., Rosnbrick, Z., Wen, Z., Yang, Z., and Zhao, Z.
\newblock The llama 3 herd of models, 2024.
\newblock URL \url{https://arxiv.org/abs/2407.21783}.

\bibitem[Fu et~al.(2023)Fu, Ou, Chen, Wan, Peng, and Khot]{fu2023chainofthought}
Fu, Y., Ou, L., Chen, M., Wan, Y., Peng, H., and Khot, T.
\newblock Chain-of-thought hub: A continuous effort to measure large language models' reasoning performance, 2023.

\bibitem[Ge et~al.(2024)Ge, Zhang, Liu, Zhang, Han, and Gao]{ge2024modeltell}
Ge, S., Zhang, Y., Liu, L., Zhang, M., Han, J., and Gao, J.
\newblock Model tells you what to discard: Adaptive kv cache compression for llms, 2024.
\newblock URL \url{https://arxiv.org/abs/2310.01801}.

\bibitem[Gunasekar et~al.(2023)Gunasekar, Zhang, Aneja, Mendes, Giorno, Gopi, Javaheripi, Kauffmann, de~Rosa, Saarikivi, Salim, Shah, Behl, Wang, Bubeck, Eldan, Kalai, Lee, and Li]{gunasekar2023textbooksneed}
Gunasekar, S., Zhang, Y., Aneja, J., Mendes, C. C.~T., Giorno, A.~D., Gopi, S., Javaheripi, M., Kauffmann, P., de~Rosa, G., Saarikivi, O., Salim, A., Shah, S., Behl, H.~S., Wang, X., Bubeck, S., Eldan, R., Kalai, A.~T., Lee, Y.~T., and Li, Y.
\newblock Textbooks are all you need, 2023.
\newblock URL \url{https://arxiv.org/abs/2306.11644}.

\bibitem[Hendrycks et~al.(2021)Hendrycks, Burns, Kadavath, Arora, Basart, Tang, Song, and Steinhardt]{hendrycks2021math}
Hendrycks, D., Burns, C., Kadavath, S., Arora, A., Basart, S., Tang, E., Song, D., and Steinhardt, J.
\newblock Measuring mathematical problem solving with the math dataset, 2021.
\newblock URL \url{https://arxiv.org/abs/2103.03874}.

\bibitem[Hong et~al.(2024)Hong, Dai, Xu, Mao, Li, Liu, Dong, Wang, et~al.]{hong2024flashdecoding}
Hong, K., Dai, G., Xu, J., Mao, Q., Li, X., Liu, J., Dong, Y., Wang, Y., et~al.
\newblock Flashdecoding++: Faster large language model inference with asynchronization, flat gemm optimization, and heuristics.
\newblock \emph{Proceedings of Machine Learning and Systems}, 6:\penalty0 148--161, 2024.

\bibitem[Joshi et~al.(2017)Joshi, Choi, Weld, and Zettlemoyer]{joshi2017triviaqa}
Joshi, M., Choi, E., Weld, D.~S., and Zettlemoyer, L.
\newblock Triviaqa: A large scale distantly supervised challenge dataset for reading comprehension, 2017.
\newblock URL \url{https://arxiv.org/abs/1705.03551}.

\bibitem[Kang et~al.(2024)Kang, Zhang, Kundu, Jeong, Liu, Krishna, and Zhao]{kang2024gear}
Kang, H., Zhang, Q., Kundu, S., Jeong, G., Liu, Z., Krishna, T., and Zhao, T.
\newblock Gear: An efficient kv cache compression recipe for near-lossless generative inference of llm, 2024.

\bibitem[Langley(2000)]{langley00}
Langley, P.
\newblock Crafting papers on machine learning.
\newblock In Langley, P. (ed.), \emph{Proceedings of the 17th International Conference on Machine Learning (ICML 2000)}, pp.\  1207--1216, Stanford, CA, 2000. Morgan Kaufmann.

\bibitem[Lin et~al.(2024)Lin, Tang, Yang, Zhang, Xiao, Gan, and Han]{lin2024qserve}
Lin, Y., Tang, H., Yang, S., Zhang, Z., Xiao, G., Gan, C., and Han, S.
\newblock Qserve: W4a8kv4 quantization and system co-design for efficient llm serving, 2024.
\newblock URL \url{https://arxiv.org/abs/2405.04532}.

\bibitem[Ling et~al.(2017)Ling, Yogatama, Dyer, and Blunsom]{ling2017program}
Ling, W., Yogatama, D., Dyer, C., and Blunsom, P.
\newblock Program induction by rationale generation: Learning to solve and explain algebraic word problems.
\newblock \emph{ACL}, 2017.

\bibitem[Liu et~al.(2023{\natexlab{a}})Liu, Zaharia, and Abbeel]{liu2023ring}
Liu, H., Zaharia, M., and Abbeel, P.
\newblock Ring attention with blockwise transformers for near-infinite context.
\newblock \emph{arXiv preprint arXiv:2310.01889}, 2023{\natexlab{a}}.

\bibitem[Liu et~al.(2023{\natexlab{b}})Liu, Wang, Dao, Zhou, Yuan, Song, Shrivastava, Zhang, Tian, Re, and Chen]{liu2023dejavu}
Liu, Z., Wang, J., Dao, T., Zhou, T., Yuan, B., Song, Z., Shrivastava, A., Zhang, C., Tian, Y., Re, C., and Chen, B.
\newblock Deja vu: Contextual sparsity for efficient llms at inference time, 2023{\natexlab{b}}.
\newblock URL \url{https://arxiv.org/abs/2310.17157}.

\bibitem[Liu et~al.(2024)Liu, Yuan, Jin, Zhong, Xu, Braverman, Chen, and Hu]{liu2024kivi}
Liu, Z., Yuan, J., Jin, H., Zhong, S., Xu, Z., Braverman, V., Chen, B., and Hu, X.
\newblock Kivi: A tuning-free asymmetric 2bit quantization for kv cache.
\newblock \emph{arXiv preprint arXiv:2402.02750}, 2024.

\bibitem[Milakov \& Gimelshein(2018)Milakov and Gimelshein]{milakov2018online}
Milakov, M. and Gimelshein, N.
\newblock Online normalizer calculation for softmax.
\newblock \emph{arXiv preprint arXiv:1805.02867}, 2018.

\bibitem[Paszke et~al.(2019)Paszke, Gross, Massa, Lerer, Bradbury, Chanan, Killeen, Lin, Gimelshein, Antiga, Desmaison, K{\"{o}}pf, Yang, DeVito, Raison, Tejani, Chilamkurthy, Steiner, Fang, Bai, and Chintala]{paszke2019pytorch}
Paszke, A., Gross, S., Massa, F., Lerer, A., Bradbury, J., Chanan, G., Killeen, T., Lin, Z., Gimelshein, N., Antiga, L., Desmaison, A., K{\"{o}}pf, A., Yang, E., DeVito, Z., Raison, M., Tejani, A., Chilamkurthy, S., Steiner, B., Fang, L., Bai, J., and Chintala, S.
\newblock Pytorch: An imperative style, high-performance deep learning library.
\newblock In Wallach, H.~M., Larochelle, H., Beygelzimer, A., d'Alch{\'{e}}{-}Buc, F., Fox, E.~B., and Garnett, R. (eds.), \emph{Advances in Neural Information Processing Systems 32: Annual Conference on Neural Information Processing Systems 2019, NeurIPS 2019, December 8-14, 2019, Vancouver, BC, Canada}, pp.\  8024--8035, 2019.

\bibitem[Rajput et~al.(2024)Rajput, Sheng, Owen, and Chiley]{rajput2024mixattention}
Rajput, S., Sheng, Y., Owen, S., and Chiley, V.
\newblock Inference-friendly models with mixattention, 2024.
\newblock URL \url{https://arxiv.org/abs/2409.15012}.

\bibitem[Sanovar et~al.(2024)Sanovar, Bharadwaj, Amant, Rühle, and Rajmohan]{sanovar2024leanatten}
Sanovar, R., Bharadwaj, S., Amant, R.~S., Rühle, V., and Rajmohan, S.
\newblock Lean attention: Hardware-aware scalable attention mechanism for the decode-phase of transformers, 2024.
\newblock URL \url{https://arxiv.org/abs/2405.10480}.

\bibitem[Shah et~al.(2024)Shah, Bikshandi, Zhang, Thakkar, Ramani, and Dao]{shah2024flashattention3}
Shah, J., Bikshandi, G., Zhang, Y., Thakkar, V., Ramani, P., and Dao, T.
\newblock Flashattention-3: Fast and accurate attention with asynchrony and low-precision, 2024.
\newblock URL \url{https://arxiv.org/abs/2407.08608}.

\bibitem[Shim et~al.(2017)Shim, Lee, Choi, Boo, and Sung]{NIPS2017_4e2a6330}
Shim, K., Lee, M., Choi, I., Boo, Y., and Sung, W.
\newblock Svd-softmax: Fast softmax approximation on large vocabulary neural networks.
\newblock In Guyon, I., Luxburg, U.~V., Bengio, S., Wallach, H., Fergus, R., Vishwanathan, S., and Garnett, R. (eds.), \emph{Advances in Neural Information Processing Systems}, volume~30. Curran Associates, Inc., 2017.
\newblock URL \url{https://proceedings.neurips.cc/paper_files/paper/2017/file/4e2a6330465c8ffcaa696a5a16639176-Paper.pdf}.

\bibitem[Suzgun et~al.(2022)Suzgun, Scales, Schärli, Gehrmann, Tay, Chung, Chowdhery, Le, Chi, Zhou, and Wei]{suzgun2022challenging}
Suzgun, M., Scales, N., Schärli, N., Gehrmann, S., Tay, Y., Chung, H.~W., Chowdhery, A., Le, Q.~V., Chi, E.~H., Zhou, D., and Wei, J.
\newblock Challenging big-bench tasks and whether chain-of-thought can solve them, 2022.

\bibitem[Tillet et~al.(2019)Tillet, Kung, and Cox]{Tillet2019TritonAI}
Tillet, P., Kung, H.-T., and Cox, D.~D.
\newblock Triton: an intermediate language and compiler for tiled neural network computations.
\newblock \emph{Proceedings of the 3rd ACM SIGPLAN International Workshop on Machine Learning and Programming Languages}, 2019.
\newblock URL \url{https://api.semanticscholar.org/CorpusID:184488182}.

\bibitem[Touvron et~al.(2023)Touvron, Lavril, Izacard, Martinet, Lachaux, Lacroix, Rozière, Goyal, Hambro, Azhar, Rodriguez, Joulin, Grave, and Lample]{touvron2023llama}
Touvron, H., Lavril, T., Izacard, G., Martinet, X., Lachaux, M.-A., Lacroix, T., Rozière, B., Goyal, N., Hambro, E., Azhar, F., Rodriguez, A., Joulin, A., Grave, E., and Lample, G.
\newblock Llama: Open and efficient foundation language models, 2023.
\newblock URL \url{https://arxiv.org/abs/2302.13971}.

\bibitem[Vasyltsov \& Chang(2021)Vasyltsov and Chang]{vasyltsov2021}
Vasyltsov, I. and Chang, W.
\newblock Efficient softmax approximation for deep neural networks with attention mechanism, 2021.
\newblock URL \url{https://arxiv.org/abs/2111.10770}.

\bibitem[Wolf et~al.(2019)Wolf, Debut, Sanh, Chaumond, Delangue, Moi, Cistac, Rault, Louf, Funtowicz, et~al.]{wolf2019huggingface}
Wolf, T., Debut, L., Sanh, V., Chaumond, J., Delangue, C., Moi, A., Cistac, P., Rault, T., Louf, R., Funtowicz, M., et~al.
\newblock Huggingface's transformers: State-of-the-art natural language processing.
\newblock \emph{arXiv preprint arXiv:1910.03771}, 2019.

\bibitem[Xiao et~al.(2024{\natexlab{a}})Xiao, Lin, Seznec, Wu, Demouth, and Han]{smoothQ}
Xiao, G., Lin, J., Seznec, M., Wu, H., Demouth, J., and Han, S.
\newblock Smoothquant: Accurate and efficient post-training quantization for large language models, 2024{\natexlab{a}}.
\newblock URL \url{https://arxiv.org/abs/2211.10438}.

\bibitem[Xiao et~al.(2024{\natexlab{b}})Xiao, Lin, Seznec, Wu, Demouth, and Han]{smoothquant}
Xiao, G., Lin, J., Seznec, M., Wu, H., Demouth, J., and Han, S.
\newblock Smoothquant: Accurate and efficient post-training quantization for large language models, 2024{\natexlab{b}}.
\newblock URL \url{https://arxiv.org/abs/2211.10438}.

\bibitem[Yang et~al.(2024)Yang, Yang, Hui, Zheng, Yu, Zhou, Li, Li, Liu, Huang, Dong, Wei, Lin, Tang, Wang, Yang, Tu, Zhang, Ma, Yang, Xu, Zhou, Bai, He, Lin, Dang, Lu, Chen, Yang, Li, Xue, Ni, Zhang, Wang, Peng, Men, Gao, Lin, Wang, Bai, Tan, Zhu, Li, Liu, Ge, Deng, Zhou, Ren, Zhang, Wei, Ren, Liu, Fan, Yao, Zhang, Wan, Chu, Liu, Cui, Zhang, Guo, and Fan]{yang2024qwen2technicalreport}
Yang, A., Yang, B., Hui, B., Zheng, B., Yu, B., Zhou, C., Li, C., Li, C., Liu, D., Huang, F., Dong, G., Wei, H., Lin, H., Tang, J., Wang, J., Yang, J., Tu, J., Zhang, J., Ma, J., Yang, J., Xu, J., Zhou, J., Bai, J., He, J., Lin, J., Dang, K., Lu, K., Chen, K., Yang, K., Li, M., Xue, M., Ni, N., Zhang, P., Wang, P., Peng, R., Men, R., Gao, R., Lin, R., Wang, S., Bai, S., Tan, S., Zhu, T., Li, T., Liu, T., Ge, W., Deng, X., Zhou, X., Ren, X., Zhang, X., Wei, X., Ren, X., Liu, X., Fan, Y., Yao, Y., Zhang, Y., Wan, Y., Chu, Y., Liu, Y., Cui, Z., Zhang, Z., Guo, Z., and Fan, Z.
\newblock Qwen2 technical report, 2024.
\newblock URL \url{https://arxiv.org/abs/2407.10671}.

\bibitem[Zhao et~al.(2024)Zhao, Lin, Zhu, Ye, Chen, Zheng, Ceze, Krishnamurthy, Chen, and Kasikci]{zhao2024atom}
Zhao, Y., Lin, C.-Y., Zhu, K., Ye, Z., Chen, L., Zheng, S., Ceze, L., Krishnamurthy, A., Chen, T., and Kasikci, B.
\newblock Atom: Low-bit quantization for efficient and accurate llm serving, 2024.
\newblock URL \url{https://arxiv.org/abs/2310.19102}.

\bibitem[Zhong et~al.(2017)Zhong, Xiong, and Socher]{zhongSeq2SQL2017}
Zhong, V., Xiong, C., and Socher, R.
\newblock Seq2sql: Generating structured queries from natural language using reinforcement learning.
\newblock \emph{CoRR}, abs/1709.00103, 2017.

\bibitem[{Zirui Liu} et~al.(2023){Zirui Liu}, {Jiayi Yuan}, {Hongye Jin}, {Shaochen Zhong}, {Zhaozhuo Xu}, Braverman, {Beidi Chen}, and Hu]{KIVI}
{Zirui Liu}, {Jiayi Yuan}, {Hongye Jin}, {Shaochen Zhong}, {Zhaozhuo Xu}, Braverman, V., {Beidi Chen}, and Hu, X.
\newblock Kivi : Plug-and-play 2bit kv cache quantization with streaming asymmetric quantization.
\newblock 2023.
\newblock \doi{10.13140/RG.2.2.28167.37282}.
\newblock URL \url{https://rgdoi.net/10.13140/RG.2.2.28167.37282}.

\end{thebibliography}
